\pdfoutput=1

\documentclass[11pt]{article}

\usepackage[final]{acl}

\usepackage{times}
\usepackage{latexsym}
\usepackage{amsmath, amssymb, graphicx, listings}
\usepackage{algorithm, algpseudocode}

\usepackage[T1]{fontenc}

\usepackage[utf8]{inputenc}

\usepackage{microtype}

\usepackage{inconsolata}

\usepackage{graphicx}

\usepackage{enumitem}
\usepackage{booktabs}
\usepackage{multirow}
\usepackage{pifont}
\usepackage{makecell}

\newcommand{\cmark}{\ding{51}}%
\newcommand{\xmark}{\ding{55}}%

\usepackage{graphicx}
\usepackage{booktabs}
\usepackage{siunitx} 
\usepackage{listings}
\lstdefinestyle{promptstyle}{
  basicstyle=\ttfamily\small,
  breaklines=true,
  columns=fullflexible,
  showstringspaces=false,
  keepspaces=true,
  mathescape=true
}
\lstnewenvironment{promptbox}[1]{%
  \subsection*{#1}%
  \lstset{style=promptstyle}%
}{}

\title{What Makes a Good Query? Measuring the Impact of Human-Confusing Linguistic Features on LLM Performance}

\author{William Watson\thanks{Equal Contribution} \qquad Nicole Cho\footnotemark[1] \qquad
  Sumitra Ganesh \qquad 
  Manuela Veloso \\
  J.P. Morgan AI Research\\
  \texttt{nicole.cho@jpmorgan.com}
  }

\begin{document}

\maketitle

\begin{abstract}
Large Language Model (LLM) hallucinations are usually treated as defects of the model or its decoding strategy. Drawing on classical linguistics, we argue that a query's form can also shape a listener's (and model's) response. We operationalize this insight by constructing a 17-dimension query feature vector covering clause complexity, lexical rarity, and anaphora, negation, answerability, and intention grounding, all known to affect human comprehension. Using \textbf{369,837} real-world queries, we ask: \textit{Are there certain types of queries that make hallucination more likely?}
A large-scale analysis reveals a consistent "risk landscape": certain features such as deep clause nesting and underspecification align with higher hallucination propensity. In contrast, clear intention grounding and answerability align with lower hallucination rates. Others, including domain specificity, show mixed, dataset- and model-dependent effects. Thus, these findings establish an empirically observable query-feature representation correlated with hallucination risk, paving the way for guided query rewriting and future intervention studies. 
\end{abstract}

\begin{figure}[t]
    \centering
    \includegraphics[width=0.95\linewidth]{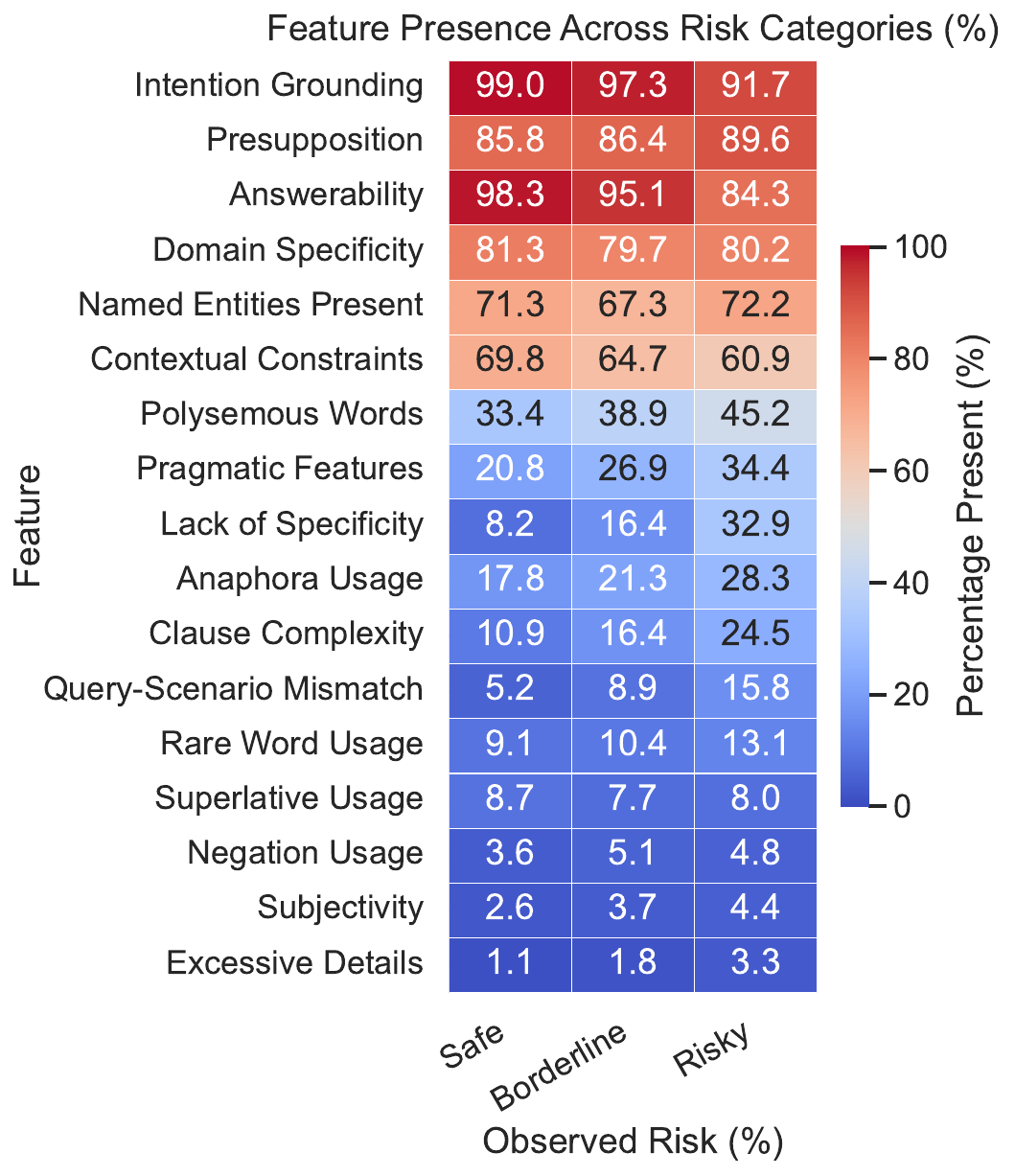}
    \vspace{-3mm}
    \caption{
    \textbf{Prevalence of binary linguistic features across hallucination risk categories (\emph{Safe}, \emph{Borderline}, \emph{Risky}).} Warmer colors indicate higher frequency. \emph{Lack of specificity}, \emph{clause complexity}, and \emph{polysemous words} show a pronounced rise from \emph{Safe} to \emph{Risky}.
    }
    \label{fig:heatmap}
    \vspace{-4.5mm}
\end{figure}

\section{Introduction}
\label{sec:into}
Large Language Models (LLMs) have transformed natural language processing, yet their propensity to hallucinate, producing plausible but factually incorrect outputs, remains a critical challenge, especially in high-stakes domains such as finance and law \citep{Huang_2025, dahl2024largelegalfictionsprofiling, naveed2024comprehensiveoverviewlargelanguage}. 
The societal, financial, and legal costs of hallucinations are already evident, with multiple lawsuits emerging in response to LLM-generated errors \citep{milmo2023lawyers}, underscoring the impracticality of relying on users to detect such failures.
While most prior work emphasizes \textit{reactive, post-generation} mitigations (e.g., self-verification, logit-based detection) \citep{lewis2021retrievalaugmentedgenerationknowledgeintensivenlp, madaan2023selfrefineiterativerefinementselffeedback} and \textit{proactive, pre-generation} strategies (e.g., RAG) \citep{lewis2021retrievalaugmentedgenerationknowledgeintensivenlp, Watson_Cho_Srishankar_2025}, comparatively fewer studies take a \emph{proactive, input-side} view beyond ambiguity detection \citep{zhang2024clamberbenchmarkidentifyingclarifying, kuhn2023clamselectiveclarificationambiguous}.

Drawing on classical linguistics, we define a 17-dimensional query feature vector capturing structural, lexical, stylistic, and semantic aspects known to shape human comprehension and obfuscate understanding. While a few of these features have been studied for their influence on general LLM performance  \citep{truong2023languagemodelsnaysayersanalysis,cho2025multiqaanalysismeasuringrobustness}, to our knowledge there is no large-scale empirical mapping from such features to \emph{hallucination} behavior. 
Following \citet{blevins2023promptinglanguagemodelslinguistic}, we leverage an LLM to extract these features from \textbf{369,837 real-world queries} spanning \textbf{13 QA datasets} (3 scenarios, 16 configurations). 

Using an semantics-preserving paraphrase neighborhood with an offline Monte Carlo correctness proxy, we provide empirical evidence of strong correlations between specific linguistic markers and observed hallucination rates, yielding a consistent "risk landscape". 
Features that \emph{destabilize} interpretation (underspecification, deep clause nesting) align with higher hallucination propensity, whereas features that \emph{tighten} semantics (clear intention grounding, answerability) align with lower risk rates. Others (e.g., domain specificity) show mixed, dataset- and model-dependent effects. 
Surprisingly, several linguistic features traditionally known to confuse human readers (e.g., word rarity, superlatives, complex negation) show minimal association with hallucination in LLMs, suggesting that human and model failure modes need not coincide. 
Our contributions are threefold:
\begin{itemize}[noitemsep, leftmargin=*, topsep=0pt, partopsep=0pt, label={\tiny\raisebox{0.5ex}{$\blacktriangleright$}}]
    \item \textbf{Feature taxonomy and extraction:} A linguistically grounded, 17-feature representation of queries known to impact language understanding for humans. We bring this perspective to LLMs and understand whether or not these features are associated with hallucinatory behavior. 
    \item \textbf{Risk landscape at scale:} An empirical, distributional map derived from ordinal modeling with dataset/scenario fixed effects and ECDF separations, linking query features to hallucination propensity over 369,837 queries. 
    \item \textbf{Proactive guidance:} We highlight practical, feature-aware triage and low-effort rewrites that can complement reactive defenses.
\end{itemize}
Therefore, we advocate a practical approach to \textit{proactively} mitigate hallucinations \textit{before} generation by optimizing queries---rather than relying on post hoc human inspection, which is often impractical.

\section{Related Work}
\label{sec:related}

\noindent
\textbf{Hallucinations in LLMs:}
Prior work addresses hallucinations both \textit{proactively} and \textit{reactively} \citep{Ji_2023, varshney2023stitchtimesavesnine, li-etal-2024-dawn}. Proactive, input-time methods (e.g., Retrieval-Augmented Generation, external tool use) enrich the context \textit{before} decoding \citep{lewis2021retrievalaugmentedgenerationknowledgeintensivenlp,schick2023toolformerlanguagemodelsteach,
qin2023toolllmfacilitatinglargelanguage}. Reactive, post-generation methods (e.g., self-consistency, logit-based detectors) evaluate or re-rank outputs after the model decodes a response \citep{wang2023selfconsistencyimproveschainthought, manakul-etal-2023-selfcheckgpt}.

\noindent
\textbf{Pre-Generation Query Evaluation:}
More recent research has begun to address hallucination proactively by examining the input query itself \citep{ji-etal-2023-towards, karpukhin-etal-2020-dense}. Studies have shown that query structure and semantic properties, such as polysemy, contextual nuance, and specificity, play a crucial role in shaping LLM outputs \citep{brown2020languagemodelsfewshotlearners, jiao2023chatgptgoodtranslatoryes}. For example, ReLA \citep{zhang2021sparseattentionlinearunits} demonstrates that sparse attention can improve both interpretability and performance without additional overhead. HalluciBot \citep{Watson_Cho_Srishankar_2025} further illustrated that perturbing queries can effectively estimate hallucination likelihood. In contrast, our work systematically extracts 22 linguistic features from queries and empirically analyzes their correlation with hallucination risk. This \textit{proactive} approach lays the foundation for query pre-filtering techniques aimed at enhancing the reliability of LLM outputs.

\section{Methodology}
\label{sec:method}

\begin{table*}[t]
    \centering
    \small
    \renewcommand{\arraystretch}{1.1}
    \setlength{\tabcolsep}{5pt}
    \begin{tabular}{l|rrrrr|rrrc}
        \toprule
        \multicolumn{1}{c}{} & \multicolumn{5}{c}{Ordinal ($\beta$-only)} & \multicolumn{4}{c}{Correlation} \\
        \cmidrule{2-10}
        \multicolumn{1}{l}{Feature} & \multicolumn{1}{c}{Coef} & \multicolumn{1}{c}{SE} & \multicolumn{1}{c}{$z$-value} & \multicolumn{1}{c}{$p$-value} & \multicolumn{1}{c|}{OR} & \multicolumn{1}{c}{$\rho$} & \multicolumn{1}{c}{$\tau$} & \multicolumn{1}{c}{Adj. $p$} & \multicolumn{1}{c}{$p<0.05$} \\
        \midrule
        \textit{Lack of Specificity}     & 0.868 & 0.010 & 85.898 & <10$^{-5}$ & 2.382 & 0.271 & 0.256 & <10$^{-5}$ & \cmark \\
        \textit{Clause Complexity}       & 0.568 & 0.010 & 57.363 & <10$^{-5}$ & 1.764 & 0.155 & 0.147 & <10$^{-5}$ & \cmark \\
        Negation Usage                   & 0.311 & 0.016 & 19.499 & <10$^{-5}$ & 1.364 & 0.028 & 0.026 & <10$^{-5}$ & \cmark \\
        Excessive Details                & 0.247 & 0.026 & 9.668  & <10$^{-5}$ & 1.281 & 0.066 & 0.063 & <10$^{-5}$ & \cmark \\
        Anaphora Usage                   & 0.214 & 0.009 & 23.827 & <10$^{-5}$ & 1.238 & 0.107 & 0.101 & <10$^{-5}$ & \cmark \\
        Polysemous Words                 & 0.096 & 0.007 & 13.840 & <10$^{-5}$ & 1.101 & 0.104 & 0.098 & <10$^{-5}$ & \cmark \\
        Rare Word Usage                  & 0.095 & 0.011 & 8.997  & <10$^{-5}$ & 1.100 & 0.055 & 0.052 & <10$^{-5}$ & \cmark \\
        Pragmatic Features               & 0.072 & 0.008 & 8.496  & <10$^{-5}$ & 1.074 & 0.132 & 0.125 & <10$^{-5}$ & \cmark \\
        Presupposition                   & 0.056 & 0.010 & 5.565  & <10$^{-5}$ & 1.058 & 0.046 & 0.044 & <10$^{-5}$ & \cmark \\
        Contextual Constraints           & 0.044 & 0.007 & 5.812  & <10$^{-5}$ & 1.045 & -0.081 & -0.077 & <10$^{-5}$ & \cmark \\
        Parse Tree Height                & 0.011 & 0.005 & 2.312  & 0.021 & 1.011 & -0.149 & -0.121 & <10$^{-5}$ & \cmark \\
        Named Entities Present           & 0.009 & 0.007 & 1.269  & 0.205 & 1.009 & 0.003 & 0.002 & 0.115      & \xmark \\
        Domain Specificity               & 0.003 & 0.009 & 0.396  & 0.692 & 1.003 & -0.013 & -0.012 & <10$^{-5}$ & \cmark \\
        Query-Scenario Mismatch          & -0.064 & 0.014 & -4.734 & <10$^{-5}$ & 0.938 & 0.153 & 0.145 & <10$^{-5}$ & \cmark \\
        Superlative Usage                & -0.103 & 0.012 & -8.674 & <10$^{-5}$ & 0.902 & -0.012 & -0.011 & <10$^{-5}$ & \cmark \\
        Dependency Depth                 & -0.128 & 0.005 & -24.353 & <10$^{-5}$ & 0.879 & -0.203 & -0.159 & <10$^{-5}$ & \cmark \\
        Intention Grounding              & -0.168 & 0.023 & -7.272 & <10$^{-5}$ & 0.846 & -0.159 & -0.151 & <10$^{-5}$ & \cmark \\
        Subjectivity                     & -0.168 & 0.019 & -8.885 & <10$^{-5}$ & 0.846 & 0.044 & 0.041 & <10$^{-5}$ & \cmark \\
        \textit{Query Token Length}      & -0.212 & 0.010 & -20.973 & <10$^{-5}$ & 0.809 & -0.274 & -0.214 & <10$^{-5}$ & \cmark \\
        \textit{Number of Clauses}       & -0.262 & 0.009 & -28.652 & <10$^{-5}$ & 0.769 & -0.272 & -0.228 & <10$^{-5}$ & \cmark \\
        \textit{Answerability}           & -1.106 & 0.017 & -63.425 & <10$^{-5}$ & 0.331 & -0.228 & -0.216 & <10$^{-5}$ & \cmark \\
        \bottomrule
    \end{tabular}
    \vspace{-1mm}
    \caption{\textbf{Results for Observed Risk analyses.} \textbf{Left:} Ordinal logistic regression estimates (using both binary and scaled numeric predictors. \textbf{Right:} Spearman's $\rho$ and Kendall's $\tau$ correlation coefficients between each feature and Observed Risk, with adjusted $p$-values and a significance indicator. Features in \textit{italics} (e.g., \textit{Lack of Specificity}, \textit{Clause Complexity}, \textit{Query Token Length}, \textit{Number of Clauses}, and \textit{Answerability}) highlight particularly intriguing effects. All adjusted $p$-values were below $10^{-5}$ except for ``Named Entities Present'' ($p=0.115$, not significant).}
    \label{tab:merged_results}
    \vspace{-3mm}
\end{table*}

\noindent
\textbf{Problem setup.}
We study how the linguistic form of a user query modulates large language model reliability. Each query $i$ receives an ordinal triage label $y_i\in\{0,1,2\}$ corresponding to
\textsc{Safe}$\,{<}\,$\textsc{Borderline}$\,{<}\,$\textsc{Risky}.
Let $x_i\in\{0,1\}^{p}$ be a binary feature vector capturing human-confusing linguistic phenomena (\S\ref{linguisticfeatures}),
and $c_i$ the observed covariates (dataset $d(i)$, scenario $s(i)$).
We model $\Pr(y_i \mid x_i,c_i)$ to quantify (i) marginal effects of features,
(ii) distributional shifts in predicted risk,
and (iii) robustness under dataset shifts--
\emph{without} rewriting queries.

\subsection{Linguistic features}
We operationalize $p{=}17$ query-level features spanning ambiguity (\textit{Lack of Specificity},
\textit{Polysemous Words}, \textit{Pragmatic Features}), referential structure (\textit{Anaphora}), complexity (\textit{Clause Complexity}), polarity (\textit{Negation}), grounding
(\textit{Answerability}, \textit{Intention Grounding}, \textit{Contextual Constraints}), and others (\S\ref{linguisticfeatures}). Detectors return structured outputs (label+rationale) via typed prompts; positive/negative
5-shot examples appear in App.~\ref{app:prompt-templates}. Detector noise is treated as classical measurement error and expected to attenuate magnitudes rather than flip signs \citep{blevins2023promptinglanguagemodelslinguistic}.

\subsection{Observed risk via semantics-preserving perturbations}
Benchmark items can be memorized, biasing raw hallucination rates \citep{carlini2021extractingtrainingdatalarge, nasr2023scalableextractiontrainingdata, aerni2024measuringnonadversarialreproductiontraining, Watson_Cho_Srishankar_2025}. For each original query $q_{\text{orig}}$, we generate a local semantic equivalence class $\mathcal{N}(q_{\text{orig}}){=}\{q_1,\dots,q_m\}$ by sampling paraphrases at $T{=}1.0$ with the instruction \textsc{"Produce a semantically indifferent but lexically perturbed version of the query."} We retain the first six paraphrases whose hybrid similarity meets $s(q_{\text{orig}},q_i)\!\ge\!0.85$,
\begin{equation*}
\begin{split}
s(q_{\mathrm{orig}}, q_i)
  &= \lambda_{\mathrm{bi}} \cdot \cos\bigl(\mathbf{e}_{\mathrm{bi}}(q_{\mathrm{orig}}), \mathbf{e}_{\mathrm{bi}}(q_i)\bigr) \\
  &\quad + \lambda_{\mathrm{cross}} \cdot \frac{1}{2} \Biggl[
      \Pr_{\mathrm{cross}}(q_{\mathrm{orig}}, q_i)\\
      &\qquad\qquad\qquad
      + \Pr_{\mathrm{cross}}(q_i, q_{\mathrm{orig}})
    \Biggr]
\end{split}
\end{equation*}
with $(\lambda_{\text{bi}},\lambda_{\text{cross}})=(0.6,0.4)$, $\mathbf e_{\text{bi}}$ from \textsc{Text-Embedding-3-Large} (3{,}072-d), and $\Pr_{\text{cross}}$ from \textsc{ms-marco-MiniLM-L6-v2} \citep{reimers-2019-sentence-bert}.

\vspace{1ex}
\noindent
\textbf{Empirical hallucination estimation.}
For each $q_i\!\in\!\mathcal{N}(q_{\text{orig}})$ we compute a convex proxy
$\hat h(q_i)=w_{0}\,s_{\text{llm}}+w_1\,s_{\text{fuzz}}+w_2\,s_{\text{bleu}}$,
combining a binary LLM-judge decision $s_{\text{llm}}\!\in\!{0,1}$ (semantic; \citealp{wang-etal-2023-chatgpt, liu2023gevalnlgevaluationusing, adlakha2024evaluatingcorrectnessfaithfulnessinstructionfollowing}), 
fuzzy string similarity $s_{\text{fuzz}}\!\in\![0,1]$ (surface; \citealp{max_bachmann_2024_10938887}),
and BLEU-1 $s_{\text{bleu}}\!\in\![0,1]$ (lexical; \citealp{Papineni02bleu:a, lin-och-2004-orange, callison-burch-etal-2006-evaluating}). 
We use $(w_0, w_1, w_2)=(0.6,0.3,0.1)$, selected on a small human-labeled set by sweeping the $(w_0', w_1', w_2')$ simplex; the ROC–AUC surface is flat for $w_0 \pm 0.2$, drops quickly with larger $w_1$, and is worst for BLEU-only, placing our mix on a Pareto plateau (App.~\ref{sec:simplex-analysis}, Fig.~\ref{app:simplex}). A perturbation counts as \emph{hallucinated} if $\hat h(q_i)>0.5$. Aggregating across the six paraphrases yields query-level categories: \textbf{Safe} (0/6), \textbf{Borderline} (1–3/6), \textbf{Risky} (4–6/6).

\begin{figure*}[t]
  \centering
  \includegraphics[width=0.93\textwidth]{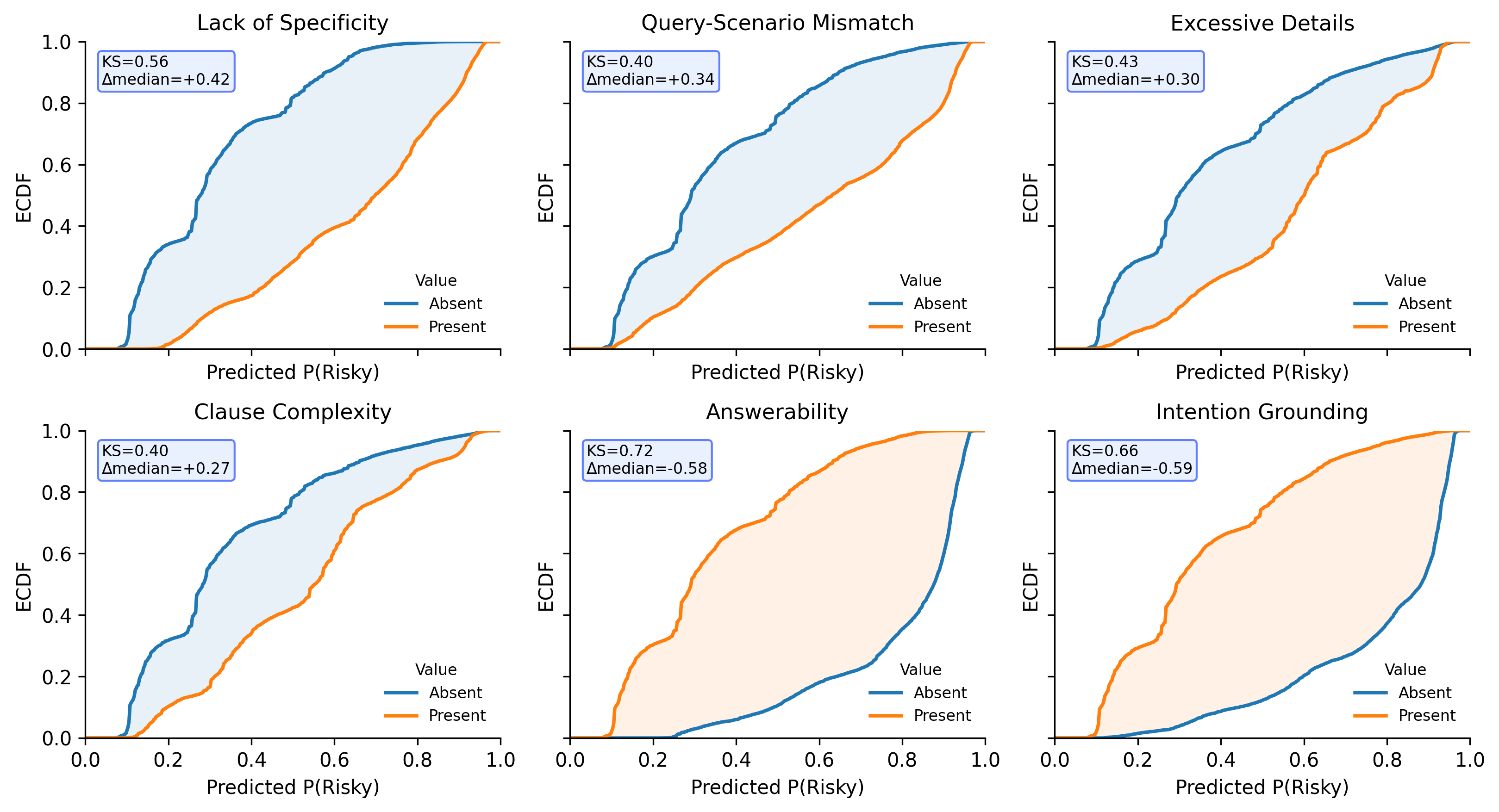}
  \caption{\textbf{ECDFs of predicted $P(\text{Risky})$ for Present vs.\ Absent (top six by KS).}
  Shaded regions indicate dominance; inset shows KS and $\Delta$median.
  Lack of Specificity, Excessive Details, Clause Complexity, and Query–Scenario Mismatch shift mass toward higher risk;
  Answerability and Intention Grounding shift mass lower.}
  \label{fig:ecdf-top6}
  \vspace{-2.0ex}
\end{figure*}

\subsection{Ordinal risk model}
\label{sec:ord-model}
We fit a proportional-odds (cumulative logit) model
\begin{equation}
\label{eq:ord}
\log\frac{\Pr(Y_i\le k\,|\,x_i,c_i)}{\Pr(Y_i>k\,|\,x_i,c_i)}=\tau_k-\eta_i
\end{equation}
with $k\in{0,1}$, linear predictor $\eta_i=x_i^\top\beta+\alpha_{d(i)}+\gamma_{s(i)}$ and ordered cutpoints $\tau_0\!<\!\tau_1$. Class probabilities are: 
\begin{align*}
p_0=&\,\,\sigma(\tau_0{-}\eta_i)\\
p_1=&\,\,\sigma(\tau_1{-}\eta_i)-\sigma(\tau_0{-}\eta_i)\\
p_2=&\,\,1-\sigma(\tau_1{-}\eta_i)
\end{align*}
optimized by NLL with $\ell_2$ penalty $\lambda_{\text{reg}}\|\beta\|_2^2$ and no explicit intercept. We report:
\begin{itemize}[noitemsep, leftmargin=*, topsep=0pt, partopsep=0pt, label={\tiny\raisebox{0.5ex}{$\blacktriangleright$}}]
\item \textbf{Specification $S_\beta$} (feature-only): $\beta$ with linguistic features only;
\item \textbf{Specification $S_{\beta, \gamma,\alpha}$} (full): $\beta$ with both scenario $\gamma$ and dataset $\alpha$ fixed effects.
\end{itemize}
Figure~\ref{fig:coef-lollipop} visualizes \emph{feature} coefficients ($\beta$; left) and \emph{dataset--scenario} effects ($\alpha, \gamma$; right). (We use $\beta$ for features throughout, reserving $\alpha$ and $\gamma$ for dataset/scenario.)

\subsection{Metrics and diagnostics}
We summarize effects at three levels:
\begin{enumerate}[noitemsep, leftmargin=*, topsep=0pt, partopsep=0pt, label={\tiny\raisebox{0.5ex}{$\blacktriangleright$}}]
\item \textbf{Coefficients} ($\beta$) from \eqref{eq:ord} under $S_\beta$ and $S_{\beta, \gamma,\alpha}$ (Fig.~\ref{fig:coef-lollipop}; Table~\ref{tab:merged_results}).
\item \textbf{Distributional separations}: ECDFs of predicted $P(\text{Risky})$ for \emph{Present} vs.\ \emph{Absent} groups; we report KS distance and $\Delta$median (Fig.~\ref{fig:ecdf-top6}).
\item \textbf{Calibration}: reliability curves and ECE within feature strata (App.~Fig.~\ref{fig:calib-all}).
\end{enumerate}
We additionally examine \textbf{length–feature interactions} by quantile-binning query length and plotting the empirical rate of a \textsc{Risky} label for \emph{Present} vs.\ \emph{Absent}, by scenario (Fig.~\ref{fig:risk-vs-length-scenario}, App.~Fig.~\ref{fig:len-grid-top}). To contextualize correlational claims, we plot \textbf{propensity overlap} (Present/Absent densities; standardized mean differences) to document where comparisons are well-posed (App.~Fig.~\ref{fig:propensity-overlap}).

\vspace{1ex}
\noindent
\textbf{Propensity modeling.}
For each binary linguistic feature $f$ (treatment $T_f\in\{0,1\}$), the \emph{propensity score} is the probability that a query exhibits $f$ given its other covariates.
Let $Z_f$ stack the remaining feature indicators $x_{-f}$ together with scenario/dataset indicators (fixed effects $\gamma,\alpha$).
We fit a separate logistic model per feature,
$\pi_f(z)\;=\;\Pr(T_f{=}1\mid Z_f{=}z)\;=\;\sigma\!\bigl(\phi_{0f}+z^\top\boldsymbol{\phi}_f\bigr),$
yielding per-item scores $\hat\pi_f=\pi_f(Z_f)$ used for overlap diagnostics.

\subsection{Robustness}
We perform Leave-One-Dataset-Out (LODO) refits of Eq.~\eqref{eq:ord} and summarize the mean $\pm$ stddev of $\beta$ across holds (Fig.~\ref{fig:lodo-stability}). Signs and relative magnitudes remain stable, indicating that the observed "risk landscape" is not driven by any single dataset.

\section{Experimental Setup}
\label{sec:experimental}

{\bf Model under test.}
All generations use \texttt{gpt-4o-2024-08-06} with a single prompting recipe held fixed across datasets; temperature $\tau=1.0$ for both answering and paraphrase sampling. Detector and audit prompts (structured outputs, 5-shot positives/negatives ICL) and sampling settings are provided in (App.~\ref{app:prompt-templates}).

\begin{figure*}[t]
  \centering
  \includegraphics[width=0.81\linewidth]{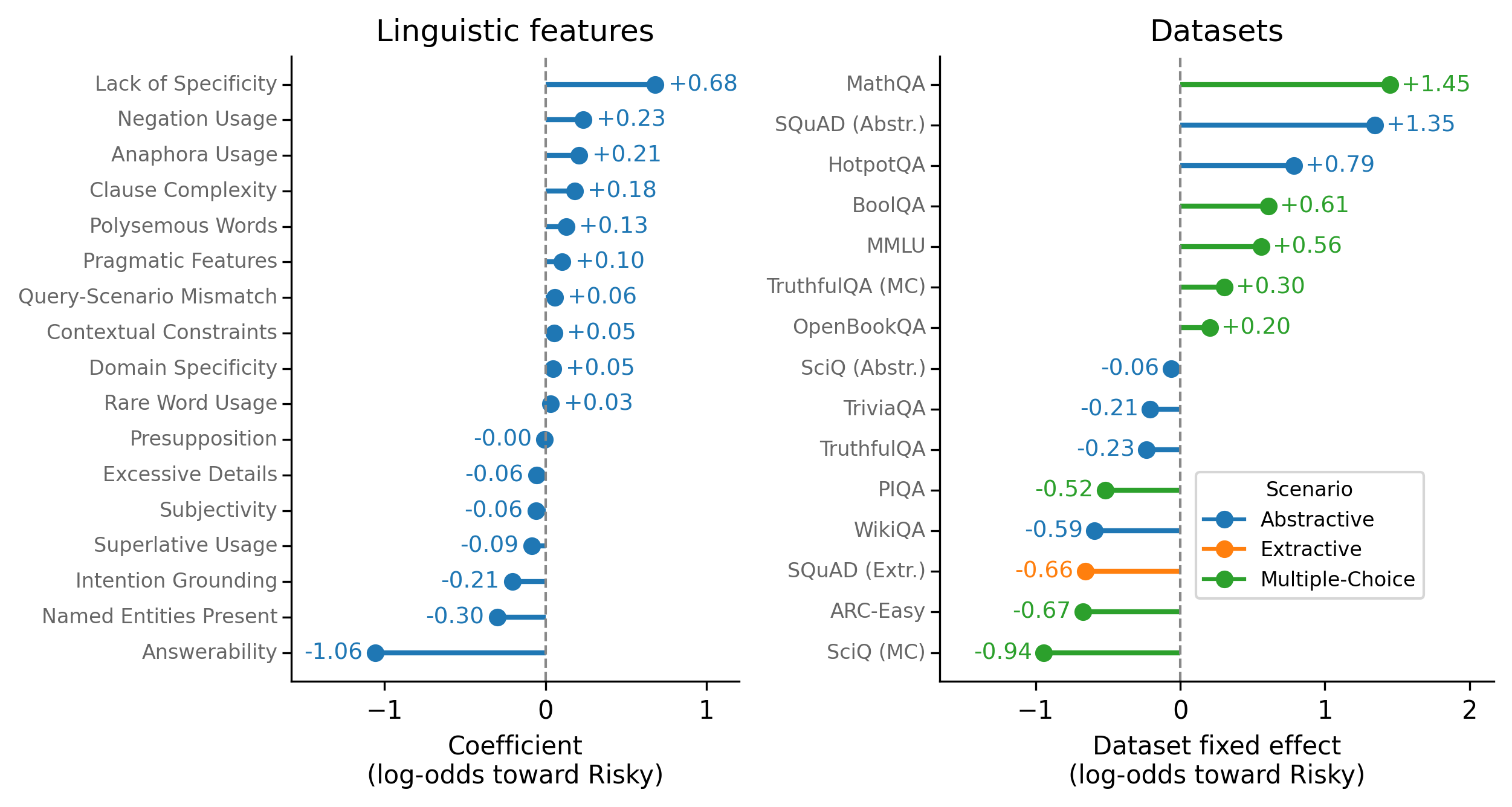}
  \caption{
  \textbf{Feature coefficients $\beta$ (left) and dataset/scenario fixed effects $\alpha,\gamma$ (right) from the ordinal logit model. Positive values increase log-odds of \emph{Risky}.}
  \emph{Answerability} is strongly protective; \emph{Lack of Specificity}, \emph{Negation}, and \emph{Anaphora} increase risk.
  }
  \label{fig:coef-lollipop}
  \vspace{-2.5ex}
\end{figure*}

\vspace{1ex}
\noindent
\textbf{Datasets and scenarios.}
We evaluate $13$ QA datasets spanning three scenarios ($16$ total configurations; Table~\ref{tab:dataset_details}):
\begin{itemize}[noitemsep, leftmargin=*, topsep=0pt, partopsep=0pt, label={\tiny\raisebox{0.5ex}{$\blacktriangleright$}}]
    \item \textbf{Extractive:} SQuADv2
    \item \textbf{Multiple Choice:} TruthfulQA, SciQ, MMLU, PIQA, BoolQ, OpenBookQA, MathQA, ARC-Easy, ARC-Challenge
    \item \textbf{Abstractive:} SQuADv2, TruthfulQA, SciQ, WikiQA, HotpotQA, TriviaQA
\end{itemize}
In total, we analyze $N=369{,}837$ query–response pairs. Scenario ($\gamma$) and dataset ($\alpha$) enter the ordinal model as fixed effects (\autoref{fig:coef-lollipop}).

\vspace{1ex}
\noindent
\textbf{Feature extraction.}
For each query we run structured detectors, producing $(\text{label}\in\{0,1\}, \text{rationale})$ per feature. Each detector's rubric is calibrated on a 100 sample held-out set to reduce systematic bias (App.~\ref{app:prompt-templates}).

\vspace{1ex}
\noindent
\textbf{Outcome construction.}
The triage label (\textsc{Safe}/\textsc{Borderline}/\textsc{Risky}) is derived from the paraphrase set using the convex hallucination proxy $\hat h(\cdot)>0.5$ threshold. We confirm that ordinal coefficients align with ECDF separations of predicted $P(\text{Risky})$ (Fig.~\ref{fig:ecdf-top6}). Ordinal KDE distributions per class are reported in Fig.~\ref{fig:kde-per-class}.

\vspace{1ex}
\noindent
\textbf{Training details.}
We implement the ordinal model in PyTorch ($1\times$NVIDIA T4), optimize NLL with Adam optimizer and $\ell_2$ regularization,
and use early stopping on a validation split \citep{kingma2017adammethodstochasticoptimization}. We fit a pooled model once and then run LODO refits (one dataset held out at a time).

\section{Results: A Query-Feature Risk Landscape for Hallucination}
\label{sec:results}

\begin{figure}[t]
  \centering
  \includegraphics[width=\linewidth]{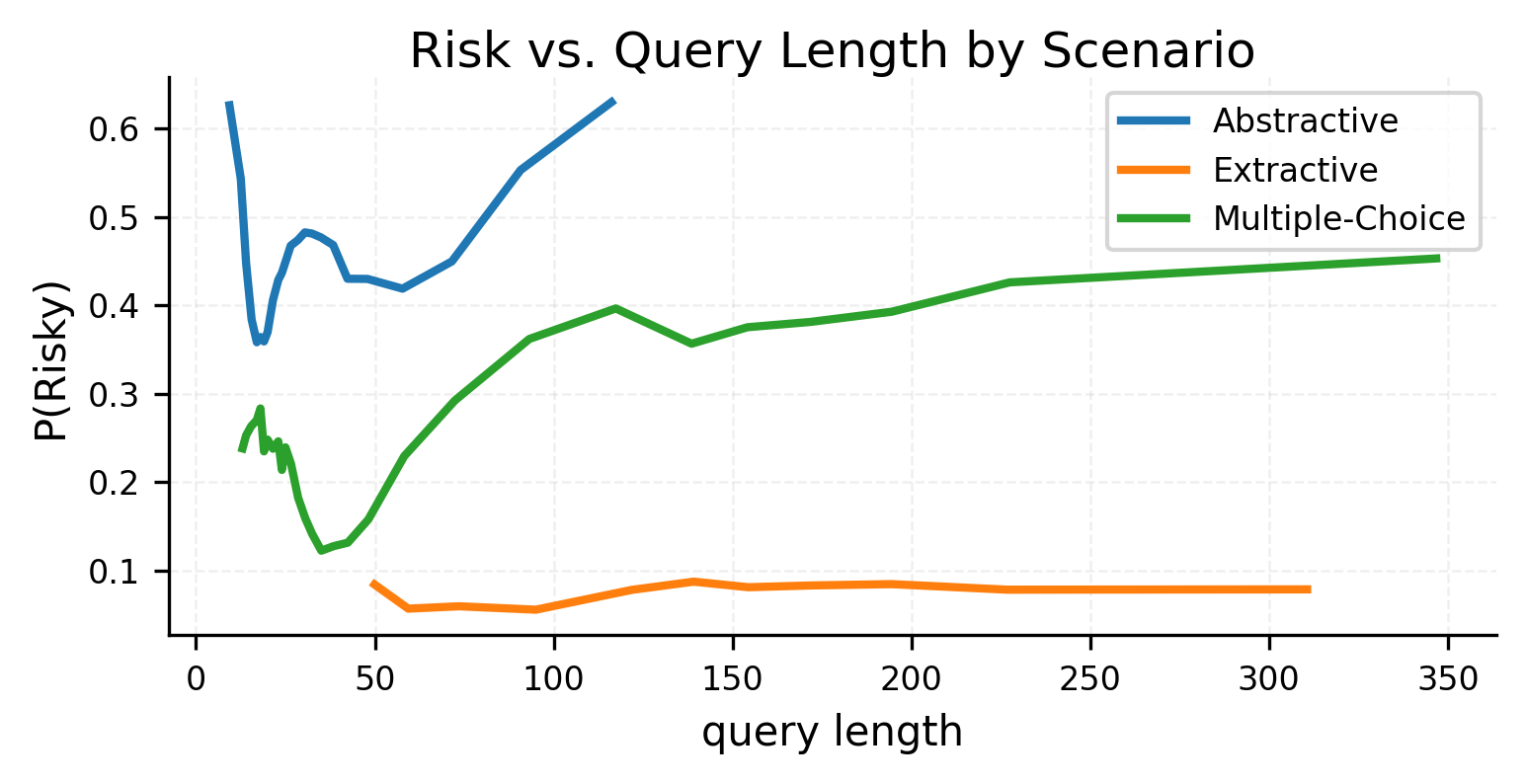}
  \caption{\textbf{Risk vs.\ query length by scenario.}
  Each curve shows the \emph{empirical} probability of a risky output (fraction of "Risky" labels) after quantile-binning query length within a scenario ($\geq50$ examples per bin).
  Risk rises with length for \textbf{Abstractive}, remains low/flat for \textbf{Extractive}, and is intermediate for \textbf{Multiple-Choice}.
  \emph{Takeaway:} longer, open-ended queries are more hallucination-prone, while extractive settings remain robust across lengths.}
  \label{fig:risk-vs-length-scenario}
  \vspace{-1.0ex}
\end{figure}

\textbf{Overview and hypotheses.}
We evaluate how human-confusing linguistic phenomena relate to LLM hallucination risk across datasets and task formats. Guided by the features in §\ref{linguisticfeatures} and the ordinal model in §\ref{sec:ord-model}, we test:
\begin{itemize}[noitemsep, leftmargin=*, topsep=0pt, partopsep=0pt, label={\tiny\raisebox{0.5ex}{$\blacktriangleright$}}]
\item \textbf{H1 (Ambiguity/complexity $\rightarrow$ higher risk):} underspecification, anaphora, negation, and clause-level complexity increase risk.
\item \textbf{H2 (Grounding $\rightarrow$ lower risk):} explicit intention and answerability reduce risk.
\item \textbf{H3 (Domain effects):} domain-specificity has mixed association, moderated by model familiarity with the domain.
\end{itemize}

\subsection{Feature and Dataset Effects}
\autoref{fig:coef-lollipop} summarizes proportional-odds estimates for two specifications: \textbf{S\textsubscript{$\beta$}} (feature-only) and \textbf{S\textsubscript{$\beta,\gamma,\alpha$}} (scenario/dataset-adjusted).
On features, \emph{Answerability} shows the largest protective effect (negative $\beta$), while \emph{Lack of Specificity}, \emph{Negation}, and \emph{Anaphora} are positively associated with risk, consistent with \textbf{H1 \& H2}. Structure-related indicators (\emph{Clause Complexity}, \emph{Polysemous Words}, \emph{Pragmatic Features}) also increase risk but with smaller magnitudes.
On contexts, fixed effects mirror scenario difficulty: abstractive configurations are riskier on average (\textsc{SQuAD (Abstr.)}, \textsc{HotpotQA}), multiple-choice safer (\textsc{SciQ}, \textsc{ARC-Easy}), and extractive in between. The signs and relative magnitudes of feature coefficients are stable with leave-one-dataset-out fits, indicating they are not artifacts of a single dataset or scenario mix.

\begin{figure}[t]
  \centering
  \includegraphics[width=\linewidth]{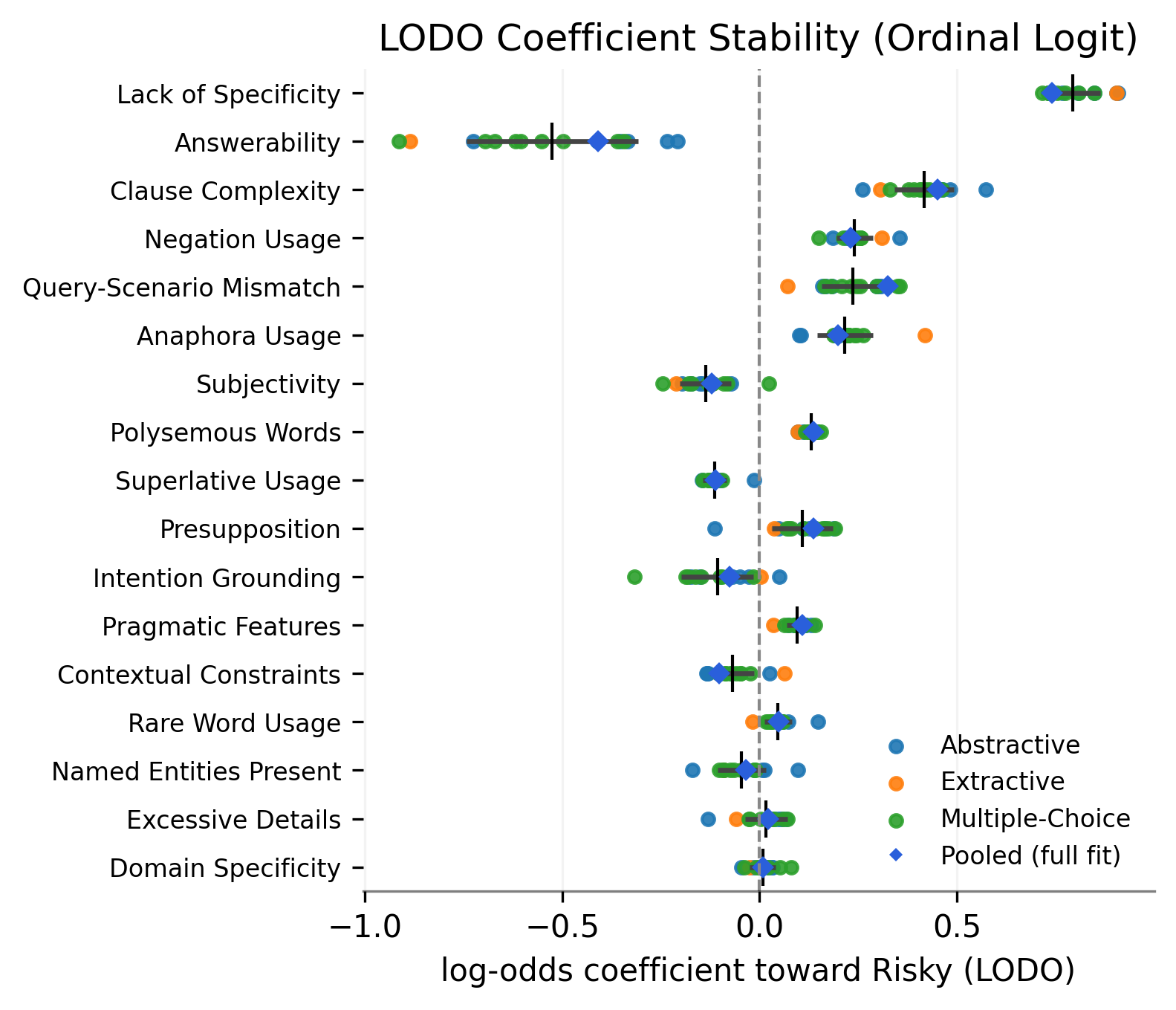}
  \caption{\textbf{LODO coefficient stability (ordinal logit).}
  Each point is a feature coefficient estimated when one dataset is held out
  (color = held-out dataset’s scenario), with short horizontal bars showing the
  mean and $\pm$1 s.d.\ across LODO runs. The blue diamond is the pooled (full-fit)
  coefficient. Signs and magnitudes are stable: \emph{Lack of Specificity},
  \emph{Clause Complexity}, \emph{Query–Scenario Mismatch} remain risk-increasing,
  while \emph{Answerability} remains strongly protective.}
  \label{fig:lodo-stability}
  \vspace{-2.0ex}
\end{figure}

\begin{figure*}[t]
    \centering
    \includegraphics[width=0.99\linewidth]{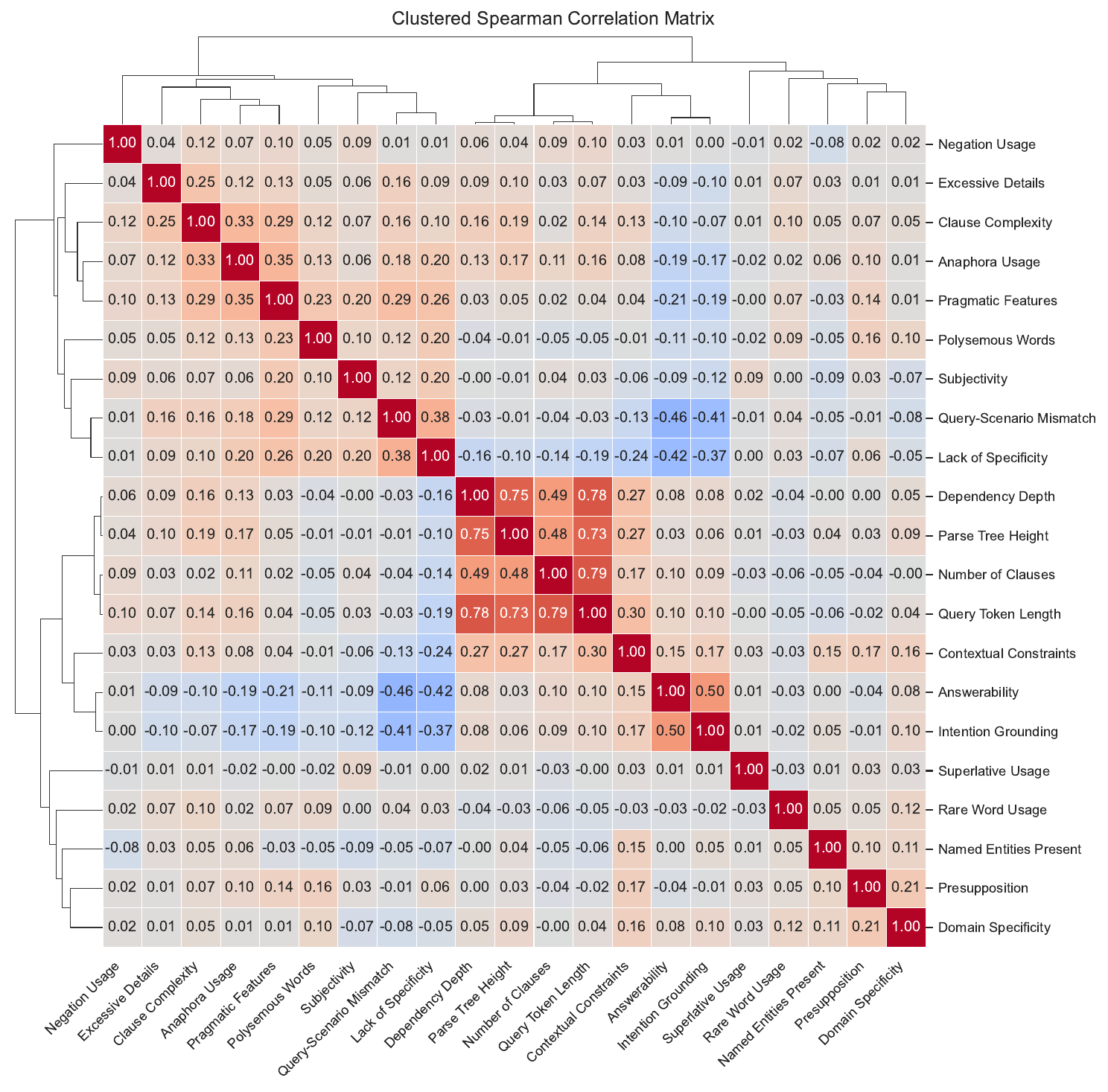}
    \vspace{-3mm}
    \caption{\textbf{Clustered Spearman correlation matrix using complete linkage \& correlation distance.} 
    The color scale ranges from red ($\rho=1$, strong positive correlation) to blue ($\rho=-1$, strong negative correlation). 
    Dendrograms group features with similar correlation patterns and share similar linguistic functions.}
    \label{fig:correlation}
    \vspace{-1.0ex}
\end{figure*}

\subsection{Distributional Effects}
To move beyond point estimates, we compare ECDFs of predicted $P(\text{Risky})$ for \emph{Present} vs.\ \emph{Absent} items per feature (\autoref{fig:ecdf-top6}). 
The top separations by KS confirm the ordinal results: \emph{Lack of Specificity}, \emph{Excessive Details}, \emph{Clause Complexity}, and \emph{Query–Scenario Mismatch} shift mass toward higher risk (positive $\Delta$median), while \emph{Answerability} and \emph{Intention Grounding} shift mass lower (negative $\Delta$median).

\subsection{Task Format Moderates Absolute Risk But Not Direction}
Baseline differences by dataset/scenario (\autoref{fig:coef-lollipop}, \autoref{fig:risk-vs-length-scenario}) are substantial:risk rises sharply with length for \textbf{Abstractive}, remains low/flat for \textbf{Extractive}, and is intermediate for \textbf{Multiple-Choice}.
Nevertheless, the \emph{direction} of feature effects are stable across bins. 
Largest gaps appear for shorter, open-ended prompts, where ambiguity features notably raise empirical \textsc{Risky} rates and grounding features reduce them. Risk–length profiles (\autoref{fig:len-grid-top}) further clarify that open-ended, longer prompts in \textbf{Abstractive} settings amplify risk, whereas \textbf{Extractive} settings remain comparatively flat across context lengths.

\subsection{Propensity overlap \& uplifts}
We estimate per-feature propensities \(\hat\pi_f=\Pr(T_f{=}1\mid Z_f)\) and plot Present/Absent KDEs to assess common support (App.\,Fig.~\ref{fig:propensity-overlap}). Where overlap is adequate, we compute uplifts in \(\Pr(\textsc{Risky})\) via IPW and stratified matching (App.\,Table~\ref{tab:uplift}).
\begin{itemize}[noitemsep, leftmargin=*, topsep=0pt, partopsep=0pt, label={\tiny\raisebox{0.5ex}{$\blacktriangleright$}}]
\item \textbf{Well-supported (uplifts reported):} \emph{Lack of Specificity}, \emph{Clause Complexity} (top-two).
  \item \textbf{Degenerate overlap (associational only):} \emph{Answerability}, \emph{Intention Grounding} (top-two).
\end{itemize}
When queries are otherwise comparable, tightening specificity and simplifying clause structure offers the clearest, overlap-supported path to reduce \(\Pr(\textsc{Risky})\); strongly protective signals like \emph{Answerability} and \emph{Intention Grounding} remain robust correlates but cannot be treated as causal toggles due to limited overlap.

\subsection{Robustness Across Datasets}
Leave-one-dataset-out refits (\autoref{fig:lodo-stability}) preserve the signs and relative ranks of the dominant features: \emph{Answerability} remains protective; \emph{Lack of Specificity}, \emph{Clause Complexity}, and \emph{Query–Scenario Mismatch} remain risk-increasing—indicating conclusions are not driven by any single dataset. Calibration within Present/Absent strata (App.Fig.\ref{fig:calib-all}) is near-diagonal with small ECE, supporting the use of probability shifts as meaningful rather than artifacts of miscalibration. Correlation structure among features (\autoref{fig:correlation}) clusters ambiguity markers together and grounding markers together, aligning with the observed risk directions.

\subsection{Linguistic Trends (Figure~\ref{fig:heatmap})}

\vspace{1ex}
\noindent
\textbf{Higher‐Risk Queries Are Marked by Ambiguity and Complexity.}
Features such as \emph{lack of specificity}, \emph{anaphora usage}, \emph{polysemy}, \emph{pragmatic features}, and \emph{clause complexity} show increased prevalence when moving from \emph{Safe} to \emph{Risky} queries, 
suggesting that higher ambiguity \& syntactic depth are more frequently associated with hallucination. 

\vspace{1ex}
\noindent
\textbf{Presupposition Is Common Across Categories.}
Interestingly, \emph{presupposition} occurs frequently even in \emph{Safe} queries, suggesting that its presence alone does not imply elevated risk. However, its co-occurrence with other risk-associated features, such as structural complexity or misaligned context, may contribute to increased hallucination rates.

\noindent
\textbf{Core Anchors of ``Safe'' Queries: Intention Grounding and Answerability.}
Queries that explicitly convey user intent (\emph{intention grounding}) and are demonstrably answerable from available context (\emph{answerability}) are highly concentrated in the \emph{Safe} category (over 90\%). This pattern aligns with the hypothesis that semantic clarity and contextual grounding are predictive of lower hallucination propensity.

\subsection{Correlation Clusters (Figure~\ref{fig:correlation})}

\vspace{1ex}
\noindent
\textbf{Syntactic Complexity:}
\emph{Query Token Length}, \emph{Dependency Depth}, \emph{Parse Tree Height}, and \emph{Number of Clauses} cluster tightly (with correlations up to $\rho=0.79$). Notably, these features exhibit significant \textbf{inverse associations} with hallucination, i.e., richer contextual cues coincide with lower risk.

\vspace{1ex}
\noindent
\textbf{Semantic Grounding:}
\emph{Intention Grounding} and \emph{Answerability} correlate strongly ($\rho=0.60$) and are moderately associated with \emph{Contextual Constraints}. This cluster is linked to lower hallucination, consistent with the hypothesis that semantically grounded queries tend to yield more accurate responses.

\vspace{1ex}
\noindent
\textbf{Ambiguity:}
\emph{Lack of Specificity}, \emph{Query-Scenario Mismatch}, \emph{Polysemous Words}, and \emph{Pragmatic Features} show moderate intercorrelations ($\rho=0.38$ between \emph{Lack of Specificity} and \emph{Query-Scenario Mismatch}). This group appears frequently in queries with higher hallucination propensity, indicating shared ambiguity-related characteristics.

\vspace{1ex}
\noindent
\textbf{Lexical and Stylistic Features:}
Attributes such as \emph{Negation Usage}, \emph{Excessive Details}, \emph{Subjectivity}, and \emph{Superlative Usage} exhibit weak correlations overall. However, these features may interact with others to influence model behavior, though their individual contributions appear limited.

\vspace{1ex}
\noindent
\textbf{Domain-Oriented Group:}
\emph{Domain Specificity}, \emph{Named Entities Present}, and \emph{Presupposition} form a loose cluster ($\rho=0.21$ for \emph{Named Entities Present} and \emph{Domain Specificity}). This suggest that domain-driven queries may entail presuppositional assumptions, which could correlate with hallucination risk when the model lacks sufficient domain familiarity.

\subsection{Regression-Based Associations with Risk}
Table~\ref{tab:merged_results} integrates our ordinal logistic regression estimates with nonparametric correlation metrics with respect to the observed hallucination rates. A \textbf{positive} coefficient indicates a feature that is positively associated with hallucination propensity, whereas a \textbf{negative} coefficient signifies an inverse association.

\noindent
\textbf{High‐Impact, Risk-Increasing:}
\begin{itemize}[noitemsep, leftmargin=*, topsep=0pt, partopsep=0pt, label={\tiny\raisebox{0.5ex}{$\blacktriangleright$}}]
    \item \emph{Lack of Specificity} presents the highest positive coefficient (0.868) and an odds ratio (OR) of 2.382, suggesting that queries which omit concrete details or precise aims are more likely to be associated with higher‐risk outputs.
    \item \emph{Clause Complexity} (0.568, OR=1.764) is also strongly associated with hallucination, consistent with the observation that syntactically intricate prompts co-occur with elevated error rates, consistent with its ECDF right-shifts.
\end{itemize}

\noindent
\textbf{Protective Features:}
\begin{itemize}[noitemsep, leftmargin=*, topsep=0pt, partopsep=0pt, label={\tiny\raisebox{0.5ex}{$\blacktriangleright$}}]
    \item \emph{Answerability} exhibits the largest negative coefficient (-1.106, OR = 0.331), suggesting that queries with clear, retrievable answers tend to have lower hallucination scores.
    \item \emph{Intention Grounding} (-0.168) is also negatively associated, indicating that queries with explicit intent are less likely to exhibit hallucination. Both align with strong left-shifts in $P(\text{Risky})$ ECDFs.
    \item Syntactic features (\emph{Query Token Length} (-0.212), \emph{Dependency Depth} (-0.128), and \emph{Number of Clauses} (-0.262)) are inversely correlated with risk, potentially reflecting that greater syntactic structure can provide helpful context.
\end{itemize}

\noindent
\textbf{Moderately Associated Features:}
\begin{itemize}[noitemsep, leftmargin=*, topsep=0pt, partopsep=0pt, label={\tiny\raisebox{0.5ex}{$\blacktriangleright$}}]
    \item \emph{Negation Usage} (0.311) and \emph{Anaphora Usage} (0.214) are positively associated with hallucination risk, possibly due to the interpretive ambiguity they introduce. However, they are both weaker than ambiguity/structure features.
    \item \emph{Polysemous Words} (0.096) broaden interpretive pathways, causing LLMs to fill gaps with hallucinated details and erroneous responses.
\end{itemize}

\noindent
\textbf{Mixed or Context-Moderated:}
\begin{itemize}[noitemsep, leftmargin=*, topsep=0pt, partopsep=0pt, label={\tiny\raisebox{0.5ex}{$\blacktriangleright$}}]
    \item \emph{Named Entities Present} is not statistically significant ($p=0.205$), no clear association between entity presence and hallucination propensity.
    \item \emph{Domain Specificity} has a near-zero coefficient (0.003, $\rho$ = -0.013), suggesting highly variable associations, possibly dependent on the model’s familiarity with the domain in question.
\end{itemize}

\subsection{Findings with respect to hypotheses}
\begin{itemize}[noitemsep, leftmargin=*, topsep=0pt, partopsep=0pt, label={\tiny\raisebox{0.5ex}{$\blacktriangleright$}}]
    \item \textbf{H1:} \emph{Lack of Specificity}, \emph{Clause Complexity}, \emph{Negation}, and \emph{Anaphora} show \textbf{strong positive associations with} hallucination risk and upward ECDF shifts.
    \item \textbf{H2:} \emph{Answerability}, \emph{Intention Grounding} exhibit substantial negative coefficients and downward ECDF shifts. This suggests that well-defined queries provide a \textbf{protective effect} against hallucinations.
    \item \textbf{H3:} \emph{Domain Specificity} has \textbf{mixed, variable associations}; effects appear moderated by dataset or model familiarity rather than uniformly positive or negative.
\end{itemize}

\subsection{Practical Applications: Risk Triage and Low-effort Rewrites}
\textbf{Triage:} At inference time, systems can (i) detect features, (ii) compute predicted $P(\text{Risky})$ under S\textsubscript{$\beta,\gamma,\alpha$}, and (iii) route high-risk queries to either a clarifying step or a retrieval/tool-grounded path.

\noindent
\textbf{Low-effort rewrites:} Our results yield three low-effort rules, directly tied to the highest-leverage features, generalize across tasks:
\begin{enumerate}[noitemsep, leftmargin=*, topsep=0pt, partopsep=0pt, label=\textbf{(\arabic*)}]
\item add disambiguating \emph{constraints} (time, place, entity) to raise specificity;
\item always state \emph{intent} explicitly (e.g., "summarize / compare / extract / verify");
\item always resolve \emph{polysemy} up front (e.g., Java the language vs.\ the island).
\end{enumerate}
Our length-conditioned profiles indicate these edits are especially important for short, open-ended prompts, where risk gaps between Present/Absent are largest. These steps are potentially automatable and align with the strongest negative coefficients and ECDF separations.


\section{Conclusion}
\label{sec:conclusion}
Taken together, our results suggest that a substantial portion of "hallucination risk" is attributable to \emph{how much the query commits the model to a determinate reading}. 
Queries that declare intent and make answerability explicit constrain the hypothesis space the model must explore; underspecified or structurally intricate queries expand that space and invite speculative completion. 
The correlation clusters are consistent with this view: grounding features co-cluster and associate with lower risk; ambiguity markers co-cluster and associate with higher risk; syntactic complexity interacts with these axes, sometimes compounding ambiguity (nested clauses, unresolved anaphora), sometimes adding helpful scaffolding when paired with explicit constraints.
These findings highlight the potential for automated query filtering and rewriting strategies to enhance model reliability by flagging risk-associated linguistic markers \textit{directly}.

\section*{Disclaimer}
{
This paper was prepared for informational purposes by the Artificial Intelligence Research group of JPMorgan Chase \& Co. and its affiliates ("JPMorgan'') and is not a product of the Research Department of JPMorgan. JPMorgan makes no representation and warranty whatsoever and disclaims all liability, for the completeness, accuracy or reliability of the information contained herein. This document is not intended as investment research or investment advice, or a recommendation, offer or solicitation for the purchase or sale of any security, financial instrument, financial product or service, or to be used in any way for evaluating the merits of participating in any transaction, and shall not constitute a solicitation under any jurisdiction or to any person, if such solicitation under such jurisdiction or to such person would be unlawful.
}

\section*{Limitations}
Our findings should be interpreted with the following limitations in mind.
First, the study is primarily observational. Although we use overlap diagnostics (propensity/positivity) and ablations to qualify comparisons, these provide, at best, quasi-causal evidence. Several features (e.g., \textit{Answerability}) are inherently semantic and not cleanly manipulable without changing meaning, so we treat their coefficients as empirical associations corroborated by multiple diagnostics rather than as causal effects.
Furthermore, our experiments are limited to English-language queries and one class of LLMs. We do not account for multimodal inputs or evolving model behavior across versions. Additionally, feature extraction relies on existing NLP toolkits and LLM predictions, which may introduce parsing errors in noisy queries.
Additionally, we treat the linguistic features as independent variables and do not model higher-order interactions. Future work could explore whether specific feature combinations jointly contribute to increased hallucination risk. Importantly, the feature correlations should not be interpreted as evidence of causality. Due to the opacity of neural representations and the challenge of tracing internal mechanisms, we frame our findings as empirical associations rather than causal claims. Finally, while our reward formulation is rigorously tuned using Pareto-optimal ROC–AUC analysis, it relies partially on an LLM-based judge, which may itself introduce systematic biases.

\bibliography{custom}

\appendix

\label{sec:appendix}

\clearpage
\clearpage

\section{Distribution of Hallucination Across Query Type}
Hallucination distributions vary across query scenarios:
\begin{itemize}[noitemsep, leftmargin=*, topsep=0pt, partopsep=0pt, label={\tiny\raisebox{0.5ex}{$\blacktriangleright$}}]
    \item \textbf{Extractive:} Hallucinations are infrequent, likely due to the presence of explicit supporting context; most queries are classified as \emph{Safe}.
    \item \textbf{Multiple Choice:} The presence of distractor options corresponds with a higher proportion of \emph{Borderline} cases.
    \item \textbf{Abstractive:} Lacking external context, abstractive queries are most frequently associated with hallucinations, with a large share labeled \emph{Risky}.
\end{itemize}

\section{Linguistic Features}
\label{linguisticfeatures}
\subsection{Structural Features}
\noindent
\textbf{Query and Context Length:}
Sentence length has long been studied as a core factor affecting  working memory performance in children \citep{Marton2006}. An interesting finding posits that syntactic complexity has a far more negative impact on a child's comprehension than sentence length \citep{Marton2006}. We bring this perspective to our study and strive to understand whether LLMs are impacted by query length. Prior studies have dived into whether LLMs can actually comprehend windows that reach their nominal capacity \citep{an2024doeseffectivecontextlength}. In contrast, instead of stress-testing the LLM by reaching the model's context window, we aim to measure the correlation between the total length of the \emph{query} and \emph{context} and hallucination propensity.

\noindent
\textbf{Anaphoric References:}
Anaphora refers to words (e.g., \emph{he, she, it, this, that, these, those}) referencing previously mentioned entities, states, or actions \citep{schuster-1988-anaphoric}. For instance, in \emph{“I like ice-cream. Do you think \textbf{it} is my favorite dessert?”}, “it” is an anaphor pointing back to “ice-cream.” Anaphoric references and their effective representations for human understanding have long confounded linguists \citep{Mann1988}. Traditional NLP research has focused on coreference resolution, linking pronominal or nominal mentions to antecedents \citep{sukthanker2018anaphoracoreferenceresolutionreview}, rather than NER. We investigate whether the presence of anaphora itself is associated with LLM errors.

\noindent
\textbf{Clause Complexity:}
Syntactic complexity is known to hinder understanding \citep{MacWhinney1984, Marton2006}. We define clause complexity as the presence of multiple subordinate clauses, which introduce syntactic dependencies. We study whether clause complexity is a feature that induces hallucinatory behavior. We count subordinate clauses using spaCy's dependency parser \citep{spacy_website} and LLM-based predictions.

\noindent
\textbf{Dependency Tree Depth:}
This metric measures how many layers of syntactic dependencies a query contains. Deeper dependency trees often involve more complex resolution chains and long-term memory, as studied in cognitive science \citep{Lewis2006}. We compute dependency depth and \textbf{Parse Tree Height}, through spaCy's dependency parser \citep{spacy_website} to understand whether it influences misunderstanding by LLMs.

\begin{table}[!t]
\centering
\small
\setlength{\tabcolsep}{4pt}
\renewcommand{\arraystretch}{1.1}
\begin{tabular}{lrrrr}
\toprule
\textbf{Query Type} & \textbf{Train} & \textbf{Val} & \textbf{Test} & \textbf{Total}\\
\midrule
\textbf{Extractive}      & 80,049 & 5,843 & --      & 85,892 \\
\textbf{Multiple Choice} & 45,997 & 14,127 & 21,573 & 81,697 \\
\textbf{Abstractive}     & 176,446 & 24,521 & 1,281 & 202,248 \\
\midrule
\textbf{Overall}         & 302,492 & 44,491 & 22,854 & 369,837 \\
\bottomrule
\end{tabular}
\caption{Number of queries across \textit{Extractive}, \textit{Multiple Choice}, and \textit{Abstractive} categories, split by train, validation (Val), and test sets. Note that we make no distinction between these splits in our analysis.}
\label{tab:query_distributions}
\end{table}

\begin{table}[!t]
\centering
\small
\resizebox{\columnwidth}{!}{
\begin{tabular}{lrrrrrr}
\toprule
 &
\multicolumn{2}{c}{\textbf{Safe}} &
\multicolumn{2}{c}{\textbf{Borderline}} &
\multicolumn{2}{c}{\textbf{Risky}} \\
\cmidrule(lr){2-3} \cmidrule(lr){4-5} \cmidrule(lr){6-7}
\textbf{Query Type} & \textbf{Count} & \textbf{\%} & \textbf{Count} & \textbf{\%} & \textbf{Count} & \textbf{\%} \\
\midrule
\textbf{Extractive}       & 58{,}834 & 69.0 & 19{,}618 & 23.0 & 6{,}773 & 8.0 \\
\textbf{Multiple Choice}  & 38{,}869 & 47.0 & 24{,}711 & 29.9 & 19{,}064 & 23.1 \\
\textbf{Abstractive}      & 67{,}078 & 33.4 & 44{,}244 & 22.0 & 89{,}429 & 44.5 \\
\bottomrule
\end{tabular}
}
\caption{Observed Risk counts and row-normalized percentages across query types. Each risk group shows the count and percentage of predictions labeled as \emph{Safe}, \emph{Borderline}, or \emph{Risky}.}
\label{tab:scenario_prediction_detailed}
\end{table}

\begin{figure*}[t]
    \centering
    \includegraphics[width=0.97\linewidth,clip,trim=1.75cm 19.85cm 3.3cm 4.52cm]{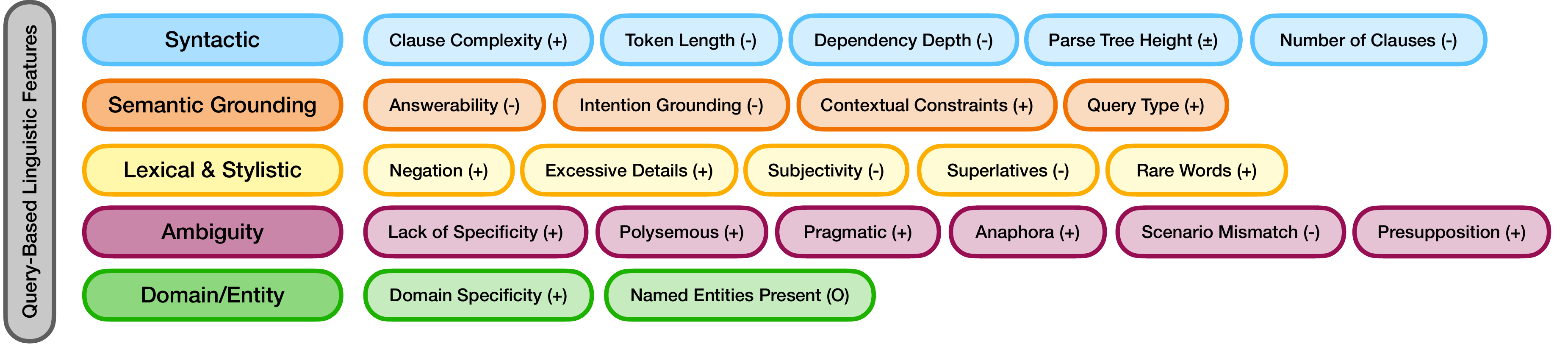}
    \vspace{-2mm}
    \caption{
    Illustration of our query-based linguistic features, grouped by correlation categories. 
    The (\textbf{+}) or (\textbf{-}) signs indicate whether each feature is positively or negatively associated with hallucination risk, 
    while “\textbf{±}” and “\textbf{O}” denote mixed or non-significant effects, respectively.
    Note: several \textbf{syntactic} features (e.g., token length, number of clauses) show \textbf{negative associations}, indicating richer structure can co-occur with \textit{lower} risk.
    }
    \label{fig:linguistic_features_diagram}
    \vspace{-4mm}
\end{figure*}

\subsection{Scenario-Based Features}

\textbf{Query Type:}
Cognitive science has long studied the different abilities required to answer open-ended (abstractive questions) compared to multiple-choice questions. \citet{Ozuru2013} studied how the efficacy in responding to open-ended questions was associated with the caliber of self-explanatory elaborations, whereas the accuracy in answering multiple-choice questions was linked to the extent of pre-existing knowledge pertinent to the text. These outcomes imply that open-ended and multiple-choice question formats assess distinct dimensions of comprehension mechanisms. While \citet{Watson_Cho_Srishankar_2025} has studied the effects of different scenarios on LLMs, we delve deeper into whether it is associated with hallucinatory outputs. 

\noindent
\textbf{Mismatch:}
Beyond query type, we also assess whether a query is aligned with the scenario in which it is posed. For instance, prompts implying an \emph{extractive} setup (\emph{“refer to recent news”}) paired with an \emph{abstractive} scenario that lacks such information. This feature is inspired by Mismatch Negativity (MMN) in cognitive science, where unexpected stimuli elicit neural responses \citep{Pulvermuller2006}. Similarly, we examine if LLMs are impacted by contextual mismatches.

\noindent
\textbf{Presupposition:}
These are implicit assumptions that a query treats as true. For example, \emph{“Who is the musician that developed neural networks?”} presupposes that such a musician exists \citep{Levinson1983}. We follow \citet{VanDerSandt1992} in identifying presuppositional triggers such as interrogative words (\emph{“Who is the football player with white hair?”}), possessive forms (\emph{“Shelley likes her dog.”}), and counterfactuals (\emph{“I would be happy if I had money.”}). Presupposition is known to hinder linguistic clarity; we study its correlation with hallucination in LLMs.

\noindent
\textbf{Pragmatics:}
Addresses context and discourse driven meanings that are not strictly encoded lexically or syntactically \citep{Levinson1983, Sadock-and-Zwicky-1985}. For instance, \emph{“Can you pass me the salt?”} is less about physical ability and more about willingness. \citet{sravanthi-etal-2024-pub} has released a benchmarks on tasks that involve understanding pragmatics - we extend this research in-depth to understand whether pragmatics impacts downstream LLM behavior.

\subsection{Lexical Features}

\noindent
\textbf{Word Rarity:}
Prior work from 2019 indicates that LLMs often struggle with rare vocabulary \citep{schick2019rarewordsmajorproblem}; as LLMs have advanced rapidly in the past few years, our motivation is to understand whether word rarity is still a risk factor for LLM understanding.

\noindent
\textbf{Negation Usage:}
Mis-primed queries including \textit{not}, \textit{never}, and \textit{no} have been shown to confuse LLMs more than humans \citep{kassner-schutze-2020-negated, truong2023languagemodelsnaysayersanalysis}.  We include negation to understand its correlation for the latest LLMs. 

\noindent
\textbf{Superlatives:}
Following \citet{scheible-2008-annotating}, superlative expressions (\textit{biggest}, \textit{fastest}, \textit{best}) indicate comparisons within a set of options that may be ambiguous or not always apparent. The interpretation of superlative adjectives has long been a study in linguistics - we therefore select it as one of our features in this study. 

\noindent
\textbf{Polysemy:}
Polysemy, or lexical ambiguity, is where words have multiple related meanings \citep{haber-poesio-2024-polysemy}. For instance, the word \textit{"mouth"} can refer either to a bodily feature or the mouth of a river.  Polysemy presents a significant challenge for English as a Second Language (ESL) learners, as it necessitates advanced cognitive processing to discern context-dependent semantic nuances and apply appropriate interpretations within varied linguistic frameworks \citep{Crossley2017}.

\subsection{Stylistic Complexity}

\textbf{Answerability:} Sarcastic or rhetorical questions pose a greater challenge for comprehension compared to straightforward, answerable queries, as they require the interpreter to discern underlying intent, contextual cues, and implicit meanings that deviate from literal interpretations, often necessitating a nuanced understanding of social and linguistic subtleties \citep{Oraby2017}. For example, the query \emph{"Based on recent news, are investors expressing concern for Stock A?"} is composed with greater clarity than \emph{"So, do you think Stock A is going to plummet?"}. Operationally, we prompt an LLM to mark a query as \textit{answerable} if the query (i) has a single or small set of verifiable answers within the provided context/dataset, (ii) is not rhetorical/sarcastic, (iii) does not require external, time-varying facts unless supplied.

\noindent
\textbf{Excessive Details:} We examine whether queries overloaded with details influence hallucination probability. While chain-of-thought prompting \citep{sahoo2024systematicsurveypromptengineering,diao2024activepromptingchainofthoughtlarge} leverages detailed reasoning, it remains uncertain whether excessive details may instead overwhelm the model and trigger hallucinations.

\noindent
\textbf{Subjectivity:} Traditional linguistics have studied the different formulations of fact-based and subjective opinions. Therefore, we strive to understand, whether a subjective opinion formulation for an LLM engenders more hallucinatory behavior. 

\noindent
\textbf{Lack of Specificity:} Queries that are broadly phrased and lack concrete details are inherently ambiguous and open to multiple interpretations \citep{brown2020languagemodelsfewshotlearners, kim-etal-2024-aligning, liu-etal-2023-afraid}. Operationally, we prompt an LLM to mark \textit{present} if $\geq1$ of: (i) missing disambiguating constraints (time/place/entity), (ii) multiple plausible interpretations without tie-breakers, (iii) underspecified task (e.g., \textit{“tell me about X”} without scope).
Not specific queries can include \textit{“Tell me about Tesla.”}, where multiple interpretations are valid (company, car, tech, stock).
A specific query contains contextual clues to identify the scope and entity discussed:  \textit{“Summarize 2024 Q4 Tesla earnings call highlights in $\leq5$ bullets.”}

\subsection{Semantic Grounding}

\textbf{Intention Grounding:}  
A query is well-grounded in intention if its purpose is immediately clear without requiring additional context \citep{Clarke}. For example, \emph{"What are the tax implications of investing in municipal bonds in the U.S.?"} in contrast to \emph{"What happens if I invest?"}.

\noindent
\textbf{Contextual Constraints:}  
Precise constraints such as specific timeframes, locations, or conditions can guide language comprehension through optimized memory storage \citep{Marton2006}. We evaluate if contextually constrained queries are less prone to hallucination. 

\noindent
\textbf{Named Entity Presence:}  
People, organizations, and places (verifiable entities) may ground LLMs in external factual information \citep{lee2023factualityenhancedlanguagemodels}. 
We take this study deeper and understand whether the presence of named entities has any measurable  impact, if any, on downstream hallucination.

\noindent
\textbf{Domain Specificity:} Domain specificity, a concept utilized across various research programs in cognitive science, refers to cognitive abilities that are constrained in specific manners. Certain cognitive abilities are confined to a particular domain, while others extend beyond it. The difficulty lies in defining the boundaries of a domain for a given capacity, particularly because knowledge areas are not inherently segmented into distinct compartments \citep{Khalidi2023}. Therefore, we measure whether  
domain-specific terminology can influence hallucination risk, depending on the model's expertise. Studies in finance \citep{zeng2024flowmindautomaticworkflowgeneration} and law \citep{watson2024lawlegalagenticworkflows} show that LLMs perform better with domain-specific tools.

\begin{table*}[!t]
    \centering
    \small
    \resizebox{\textwidth}{!}{%
    \begin{tabular}{lcccccc}
        \toprule
        \textbf{Dataset} & \textbf{Scenario} & \textbf{Domain} & \textbf{License} & \textbf{Count} & \textbf{Citation} \\
        \midrule
        SQuADv2         & E, A   & Wikipedia              & CC BY-SA 4.0 & 86K  & \citet{rajpurkar2016squad, rajpurkar2018know} \\
        TruthfulQA      & M, A   & General Knowledge      & Apache-2.0   & 807  & \citet{lin-etal-2022-truthfulqa} \\
        SciQ            & M, A   & Science                & CC BY-NC 3.0 & 13K  & \citet{SciQ} \\
        MMLU            & M      & Various                & MIT          & 15K  & \citet{hendryckstest2021} \\
        PIQA            & M      & Physical Commonsense   & AFL-3.0      & 17K  & \citet{Bisk2020} \\
        BoolQ           & M      & Yes/No Questions       & CC BY-SA 3.0 & 13K  & \citet{clark2019boolq, wang2019superglue} \\
        OpenBookQA      & M      & Science Reasoning      & Apache-2.0   & 6K   & \citet{OpenBookQA2018} \\
        MathQA          & M      & Mathematics            & Apache-2.0   & 8K   & \citet{amini-etal-2019-mathqa} \\
        ARC-Easy        & M      & Science                & CC BY-SA 4.0 & 5K   & \citet{allenai:arc} \\
        ARC-Challenge   & M      & Science                & CC BY-SA 4.0 & 2.6K & \citet{allenai:arc} \\
        WikiQA          & A      & Wikipedia QA           & Other        & 1.5K & \citet{yang-etal-2015-wikiqa} \\
        HotpotQA        & A      & Multi-hop Reasoning    & CC BY-SA 4.0 & 72K  & \citet{yang-etal-2018-hotpotqa} \\
        TriviaQA        & A      & Trivia                 & Apache-2.0   & 88K  & \citet{2017arXivtriviaqa} \\
        \bottomrule
    \end{tabular}%
    }
    \caption{Overview of datasets used in our study, including domain, license, number of examples, and associated scenario types. These datasets span a diverse range of question types, knowledge areas, and reasoning skills, supporting robust evaluation across domains. Scenario types tested: E = Extractive, M = Multiple Choice, A = Abstractive.}
    \label{tab:dataset_details}
    \vspace{-5mm}
\end{table*}

\section{Reward-Weight Simplex Analysis:}
\label{sec:simplex-analysis}
\begin{figure}[t]
    \centering
    \includegraphics[width=0.98\columnwidth]{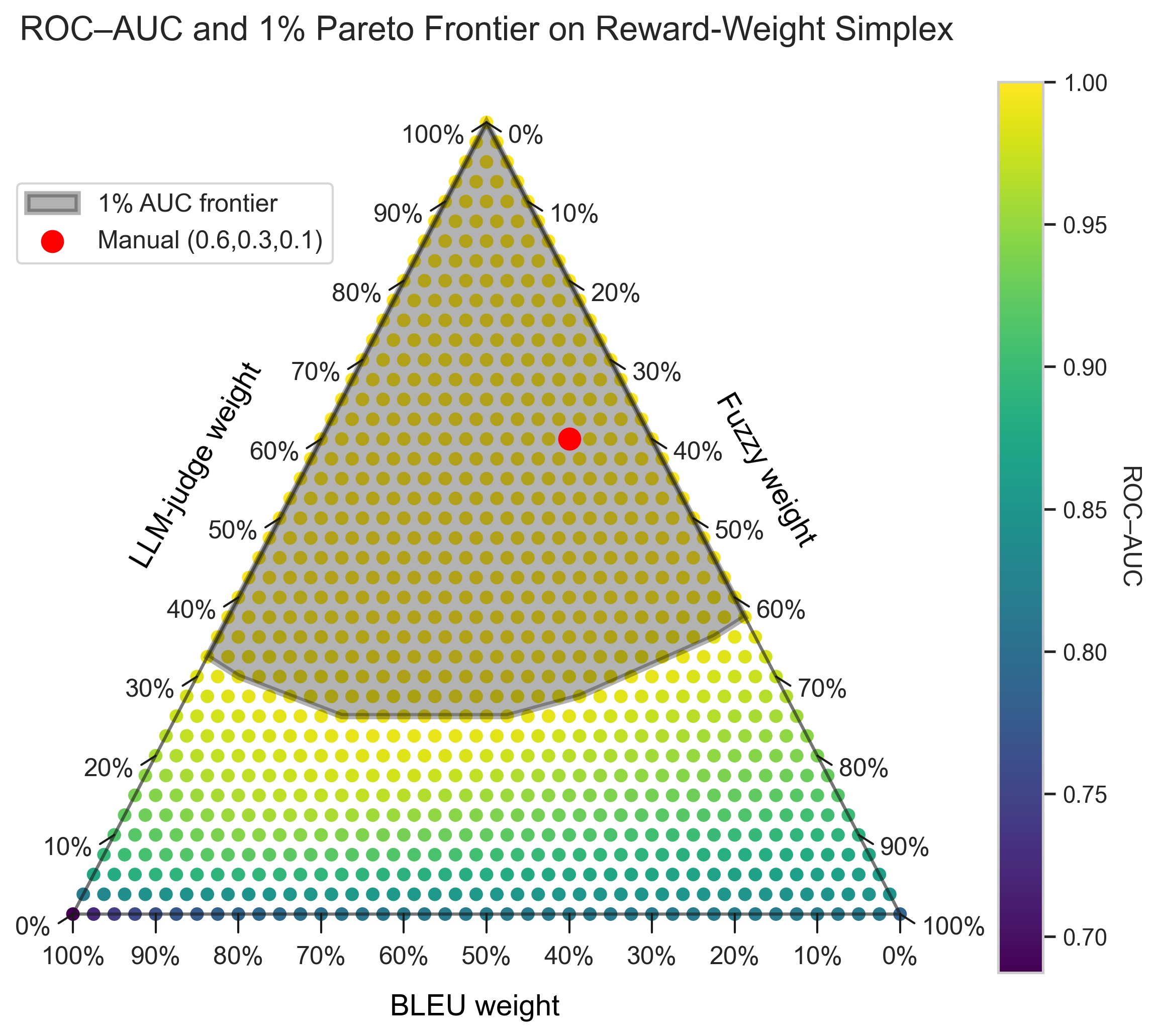}
    \caption{
        ROC--AUC Landscape over Reward-Weight Simplex.
        Each point represents a convex combination of weights \((\alpha, \beta, \gamma)\) over the LLM-judge, Fuzzy, and BLEU metrics. Color indicates ROC--AUC measured on a held-out validation set; the shaded region denotes the top 1\% frontier. Our selected weights \((0.6, 0.3, 0.1)\) are marked in red.
    }
    \label{app:simplex}
\end{figure}

We swept \((w_0',w_1',w_2')\) over a triangular grid (\(w_0'+w_1'+w_2'=1\)) and computed ROC–AUC on the 100 item human‐labeled validation set. 
AUC degrades when relying on BLEU alone, and increases when dominated by the judge; the $0.6/0.3/0.1$ convex mix is on the Pareto plateau.
The Pareto frontier (Appendix~\ref{app:simplex}) for $\hat{h}(q_i)$ reveals the following:
\begin{itemize}[noitemsep, leftmargin=*, topsep=0pt, partopsep=0pt, label={\tiny\raisebox{0.5ex}{$\blacktriangleright$}}]
  \item \textbf{LLM‐Judge Robustness (\(w_0\)):}  
    The ROC–AUC surface is nearly invariant when \(w_0\) varies by \(\pm0.2\): AUC shifts by <0.5\%, indicating our formulation tolerates large $w_0$ weight swings.
  \item \textbf{Fuzzy‐Match Sensitivity (\(w_1\)):}  
    Small increases in \(w_1\) rapidly exit the Pareto region, showing that the fuzzy‐match term must be tuned carefully to avoid degrading overall accuracy.
\item \textbf{BLEU‐Only Pitfall (\(w_2\)):}  
    As \(w_2\) increases, AUC steadily declines, bottoming out at \(w_2=1\), where the metric overemphasizes surface overlap at the expense of semantic correctness.

  \item \textbf{Pareto‐Optimal Region:}  
    We select ($0.6$, $0.3$, $0.1$) as our final weights, which lie deep in the high‐AUC plateau, confirming it is a Pareto‐optimal trade‐off among semantic, fuzzy, and lexical signals.
\end{itemize}

\section{Feature Calculation Methodology}

We computed our linguistic features using a combination of techniques:

\begin{itemize}[noitemsep, leftmargin=*, topsep=0pt, partopsep=0pt, label={\tiny\raisebox{0.5ex}{$\blacktriangleright$}}]
    \item \textbf{LLM Structured Output Model:}  
    For binary features, we employed an LLM structured output model (\texttt{gpt-4o-2024-08-06}) that leverages a Pydantic schema. This schema includes, for every feature dimension, a chain-of-thought slack variable, enabling the model to consider all relevant variables before predicting the final boolean values for each feature. Our in-context examples and definitions are itemized in Table~\ref{tab:feature_dimensions}.
    
    \item \textbf{spaCy Parsers:}  
    Syntactic features such as the number of clauses, dependency depth, and parse tree height were computed using spaCy's parsers, which provided robust dependency and constituency parsing capabilities \citep{spacy_website}.
    
    \item \textbf{OpenAI's tiktoken Library:}  
    Token lengths were determined using OpenAI's \texttt{tiktoken} library,\footnote{\url{https://github.com/openai/tiktoken}} with encoding \texttt{o200k\_base}, ensuring consistency with the tokenization process used during simulation. Note that the observed risk is derived from responses generated with \texttt{gpt-4o-2024-08-06}.
\end{itemize}

\begin{figure*}[t]
  \centering
  \includegraphics[width=\linewidth]{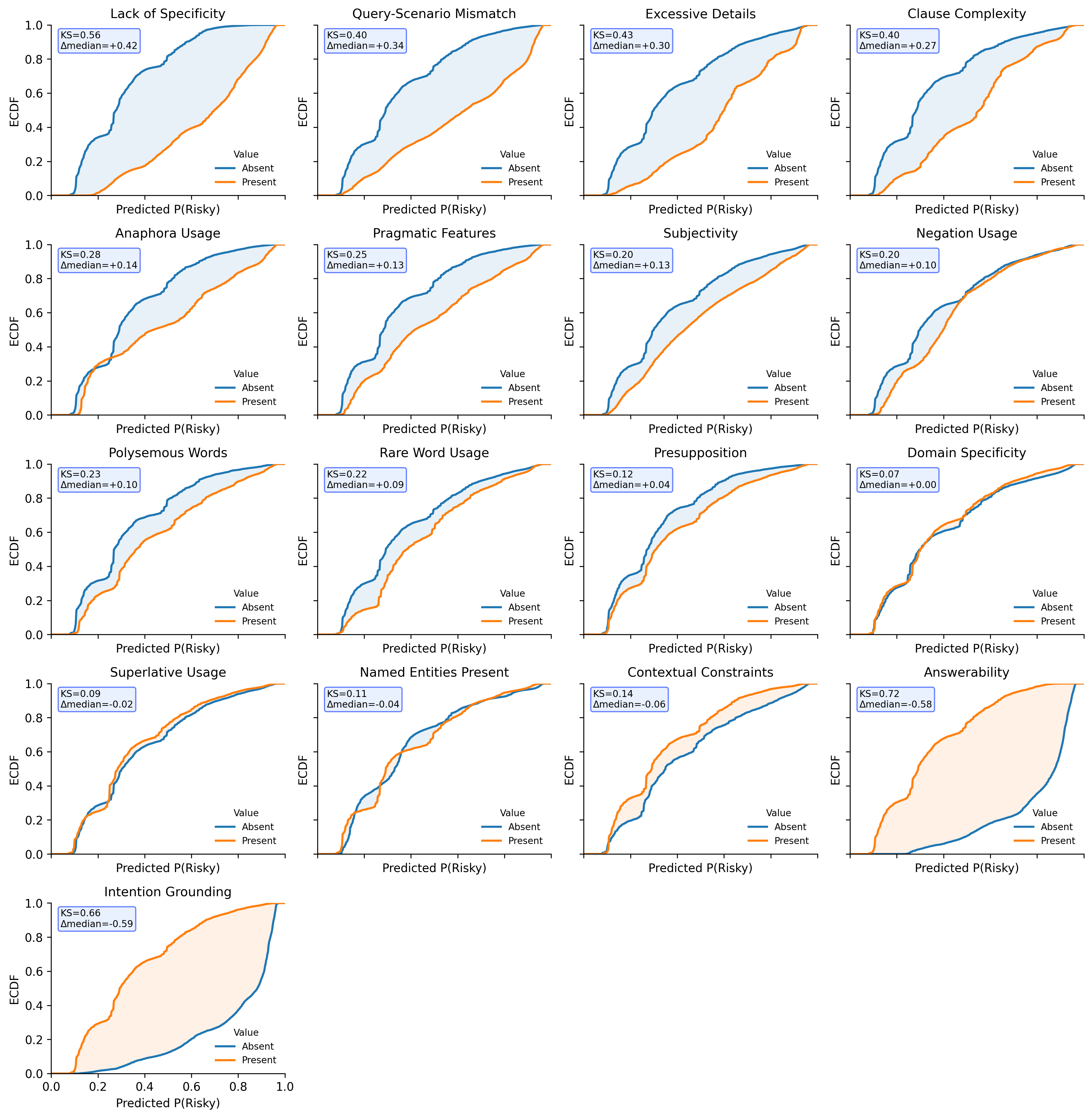}
  \caption{\textbf{All features: ECDFs of predicted $P(\text{Risky})$ by feature presence.}
  Same rendering as \autoref{fig:ecdf-top6}.}
  \label{fig:ecdf-all}
\end{figure*}

\begin{table*}[t]
  \centering
  \small
  \begin{tabular}{@{}l S[table-format=1.2] S[table-format=+1.2] r r l@{}}
    \toprule
    Feature & {KS} & {$\Delta$median} & {$n_{\text{abs}}$} & {$n_{\text{pres}}$} & Direction \\
    \midrule
    Answerability            & 0.72 & -0.58 & 25,280  & 343,340 & risk $\downarrow$ \\
    Intention Grounding      & 0.66 & -0.59 & 13,576  & 355,044 & risk $\downarrow$ \\
    Lack of Specificity      & 0.56 & +0.42 & 302,781 & 65,839  & risk $\uparrow$ \\
    Excessive Details        & 0.43 & +0.30 & 361,479 & 7,141   & risk $\uparrow$ \\
    Clause Complexity        & 0.40 & +0.27 & 307,849 & 60,771  & risk $\uparrow$ \\
    Query–Scenario Mismatch  & 0.40 & +0.34 & 333,939 & 34,681  & risk $\uparrow$ \\
    Anaphora Usage           & 0.28 & +0.14 & 287,833 & 80,787  & risk $\uparrow$ \\
    Pragmatic Features       & 0.25 & +0.13 & 270,893 & 97,727  & risk $\uparrow$ \\
    Polysemous Words         & 0.23 & +0.10 & 226,974 & 141,646 & risk $\uparrow$ \\
    Rare Word Usage          & 0.22 & +0.09 & 329,378 & 39,242  & risk $\uparrow$ \\
    Negation Usage           & 0.20 & +0.10 & 352,690 & 15,930  & risk $\uparrow$ \\
    Subjectivity             & 0.20 & +0.13 & 356,043 & 12,577  & risk $\uparrow$ \\
    Contextual Constraints   & 0.14 & -0.06 & 126,145 & 242,475 & risk $\downarrow$ \\
    Presupposition           & 0.12 & +0.04 & 47,411  & 321,209 & risk $\uparrow$ \\
    Named Entities Present   & 0.11 & -0.04 & 108,331 & 260,289 & risk $\downarrow$ \\
    Superlative Usage        & 0.09 & -0.02 & 338,308 & 30,312  & risk $\downarrow$ \\
    Domain Specificity       & 0.07 & +0.00 & 71,623  & 296,997 & risk$\uparrow$ \\
    \bottomrule
  \end{tabular}
  \caption{\textbf{Feature ranking by ECDF separation.} KS and $\Delta$median computed on predicted $P(\text{Risky})$.}
  \label{tab:ecdf-all}
\end{table*}

\begin{table*}[t]
\centering
\small
\setlength{\tabcolsep}{5pt}
\begin{tabular}{lrrrrr}
\toprule
\textbf{Feature} & \textbf{$n_{\text{Present}}$} & \textbf{$n_{\text{Absent}}$} & \textbf{Overlap} & \textbf{ATE (IPW)} & \textbf{ATE (Strat.)} \\
\midrule
Named Entities Present     & 260{,}289 & 108{,}331 & 1.000 &  -0.000 &  -0.001 \\
Polysemous Words           & 141{,}646 & 226{,}974 & 1.000 &  +0.014 &  +0.016 \\
Pragmatic Features         &  97{,}727 & 270{,}893 & 0.993 &  +0.007 &  +0.006 \\
Contextual Constraints     & 242{,}475 & 126{,}145 & 0.983 &  +0.000 &  +0.003 \\
Domain Specificity         & 296{,}997 &  71{,}623 & 0.979 &  +0.001 &  +0.006 \\
Clause Complexity          &  60{,}771 & 307{,}849 & 0.969 &  \textbf{+0.103} &  \textbf{+0.083} \\
Rare Word Usage            &  39{,}242 & 329{,}378 & 0.937 &  +0.015 &  +0.009 \\
Anaphora Usage             &  80{,}787 & 287{,}833 & 0.918 &  \textbf{+0.059} &  \textbf{+0.071} \\
Superlative Usage          &  30{,}312 & 338{,}308 & 0.890 &  \textbf{-0.032} &  \textbf{-0.025} \\
Presupposition             & 321{,}209 &  47{,}411 & 0.838 &  +0.016 &  +0.007 \\
Lack of Specificity        &  65{,}839 & 302{,}781 & 0.808 &  \textbf{+0.212} &  \textbf{+0.199} \\
Query--Scenario Mismatch   &  34{,}681 & 333{,}939 & 0.488 &  +0.039 &  +0.012 \\
Answerability              & 343{,}340 &  25{,}280 & 0.338 &    ---  &    ---  \\
Negation Usage             &  15{,}930 & 352{,}690 & 0.338 &    ---  &    ---  \\
Subjectivity               &  12{,}577 & 356{,}043 & 0.266 &    ---  &    ---  \\
Intention Grounding        & 355{,}044 &  13{,}576 & 0.225 &    ---  &    ---  \\
Excessive Details          &   7{,}141 & 361{,}479 & 0.126 &    ---  &    ---  \\
\bottomrule
\end{tabular}
\caption{\textbf{Propensity overlap and overlap-conditioned uplifts by feature.}
Overlap is the common-support share in \(\pi_f(z)\) (Present vs.~Absent).
ATEs are percentage-point changes in \(\Pr(\textsc{Risky})\) under IPW and propensity-stratified matching, reported only where overlap is adequate; \textbf{bold} marks salient non-zero effects. ``---'' indicates poor overlap (no uplift reported).}
\label{tab:uplift}
\end{table*}

\section{Ordinal Logistic Regression Details}
\label{sec:ord-model-app}

We use an ordinal logistic regression model (\texttt{OrderedModel} from \texttt{statsmodels} \citep{seabold2010statsmodels}) to estimate the effect of linguistic features on hallucination risk. The response variable is ordinal with three levels: \textit{Safe}, \textit{Borderline}, and \textit{Risky}. Predictor variables include the binary linguistic features described in Section~\ref{sec:method}.
Trained using the BFGS optimization method. Table~\ref{tab:merged_results} reports the estimated coefficients, where positive values indicate a higher likelihood of hallucination risk.

\paragraph{Analysis of Results.}
For each binary feature $f$, we plot ECDFs of model-predicted $P(\text{Risky})$ for $f{=}0$ vs $f{=}1$, report the Kolmogorov–Smirnov distance (KS) and $\Delta$median = $\mathrm{median}(P(\text{Risky}) \mid f{=}1) - \mathrm{median}(P(\text{Risky}) \mid f{=}0)$, and shade the region of dominance.
For length analyses we partition queries into $Q{=}30$ equal-mass bins by token length.
Within each bin we compute the empirical rate of the \emph{Risky} label for items
with a feature \emph{Present} vs \emph{Absent}. We plot the bin means against the
bin centers, with binomial 95\% confidence bands. For ECDFs we compare the
distributions of model-predicted $P(\text{Risky})$ under Present vs Absent and
report the Kolmogorov–Smirnov distance and $\Delta$median. The distributions exhibit the expected ordering (e.g., \emph{Risky} items have higher $P(\text{Risky})$ mass). 

\paragraph{Calibration.}
We bucket predicted $P(\text{Risky})$ into 10 equal-mass bins and plot observed
frequency vs mean predicted probability, separately for Present/Absent per feature.
Expected Calibration Error (ECE) is the weighted average of absolute deviations
between observed and predicted bin rates.

\section{Propensity and IPW Uplift Computation}
\label{app:ipw}

\textbf{Setup.}
For each binary linguistic feature $f$ (treatment $T_{fi}\!\in\!\{0,1\}$ for item $i$), let $Z_{fi}$ stack all \emph{other} feature indicators $x_{-f,i}$ together with scenario and dataset indicators ($\gamma,\alpha$). The outcome used for uplift is a scalar $O_i\!\in\![0,1]$ (either the observed risky label $\mathbf{1}\{y_i{=}\textsc{Risky}\}$ or the model-implied $P_i(\textsc{Risky})$).

\noindent
\textbf{Propensity model.}
We estimate the feature-specific propensity
\[
\pi_f(z)\;=\;\Pr(T_f{=}1\mid Z_f{=}z)
\]
with a separate logistic regression per $f$:
\[
\hat\pi_{fi}\;=\;\sigma\!\bigl(\phi_{0f}+Z_{fi}^{\!\top}\boldsymbol{\phi}_f\bigr),
\]
holding the global covariate set fixed but excluding $f$ to avoid leakage. To prevent extreme weights, we bound $\hat\pi_{fi}\!\in\![10^{-3},1{-}10^{-3}]$.

\begin{figure*}[t]
  \centering
  \includegraphics[width=0.9\linewidth]{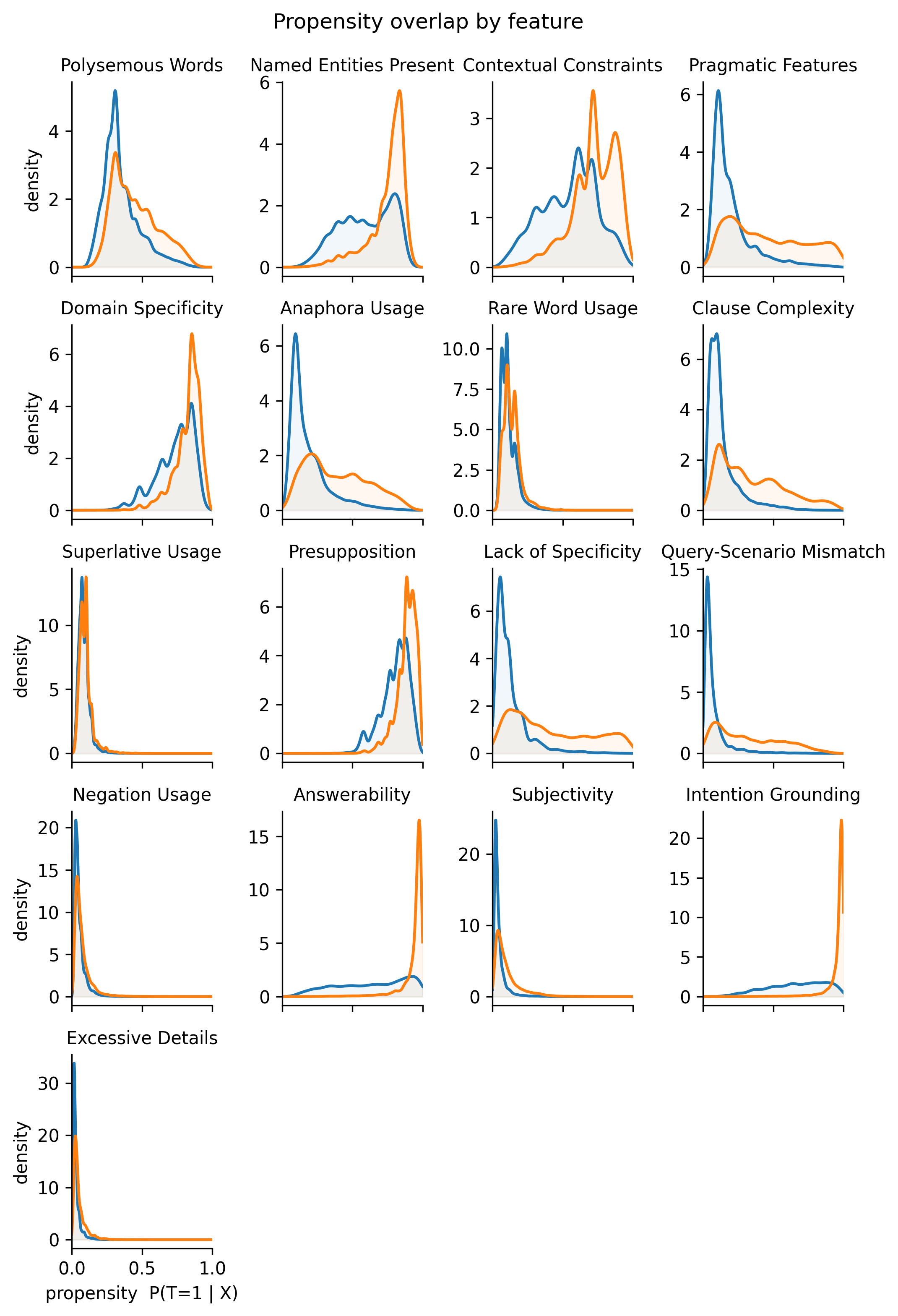}
  \caption{\textbf{Propensity overlap by feature.}
  For each feature $f$, we fit a logistic model
$\pi_f(z)=\Pr(T_f{=}1 \mid Z_f{=}z)$ over $Z_f=(x_{-f},\alpha,\gamma)$ and plot KDEs of $\hat\pi_f$ for \emph{Present} ($T_f{=}1$) vs.\ \emph{Absent} ($T_f{=}0$).
Substantial
  overlap indicates adequate support for balancing or matching; near-degenerate
  propensities (mass near 0 or 1) warn that causal comparisons will be fragile.
  Several features (e.g., \emph{Answerability}, \emph{Intention Grounding})
  show limited overlap, which we treat with weighting and sensitivity analyses.}
  \label{fig:propensity-overlap}
\end{figure*}

\noindent
\textbf{Overlap (positivity) diagnostic.}
We quantify common support via the \emph{overlap share}
\[
\mathrm{overlap}_f \;=\; \frac{1}{n}\sum_{i=1}^n 
\mathbf{1}\!\bigl\{\hat\pi_{fi}\!\in[\alpha,1{-}\alpha]\bigr\}
\]
where $\alpha{=}0.05$, and visualize $\hat\pi_{fi}$ for \emph{Present} vs.\ \emph{Absent} with KDEs (Fig.~\ref{fig:propensity-overlap}). We report matched/Inverse Probability Weighting (IPW) uplifts only where overlap is substantial (see App.~Table~\ref{tab:uplift}).

\paragraph{IPW uplift.}
The inverse-probability-weighted (IPW) ATE for feature $f$ on $O$ is
\[
\hat\tau^{\text{IPW}}_f \;=\;
\frac{\sum_i \frac{T_{fi}}{\hat\pi_{fi}}\,O_i}{\sum_i \frac{T_{fi}}{\hat\pi_{fi}}}
\;-\;
\frac{\sum_i \frac{1{-}T_{fi}}{1{-}\hat\pi_{fi}}\,O_i}{\sum_i \frac{1{-}T_{fi}}{1{-}\hat\pi_{fi}}}.
\]
IPW is interpretable as a quasi-causal contrast under unconfoundedness and positivity; where overlap is weak, we treat estimates as associational.

\noindent
\textbf{Matched (stratified) contrast.}
As a complementary, low-variance estimator, we stratify by $\hat\pi_{fi}$ quantiles into $K{=}10$ bins $b$ and compute
\[
\hat\tau^{\text{match}}_f \;=\; \sum_{b=1}^{K} \omega_b\!\left(\,\bar O_{1b}\;-\;\bar O_{0b}\right)
\]
where $\omega_b \propto n_b$ and $\bar O_{tb}$ is the within-bin mean outcome for $T{=}t$ and $n_b$ is the bin size. We report both $\hat\tau^{\text{IPW}}_f$ and $\hat\tau^{\text{match}}_f$ when overlap is adequate (App.~Table~\ref{tab:uplift}).

\noindent
\textbf{Interpretation.}
These contrasts estimate the change in risky outcome associated with \emph{feature presence}, conditional on the other query features and dataset/scenario mix. Practically, we trust uplift magnitudes only for features with strong overlap; for near-degenerate features (e.g., \emph{Answerability}, \emph{Intention Grounding}), we report coefficients and ECDF gaps as correlational signals.

\begin{figure*}[t]
  \centering
  \includegraphics[width=\linewidth]{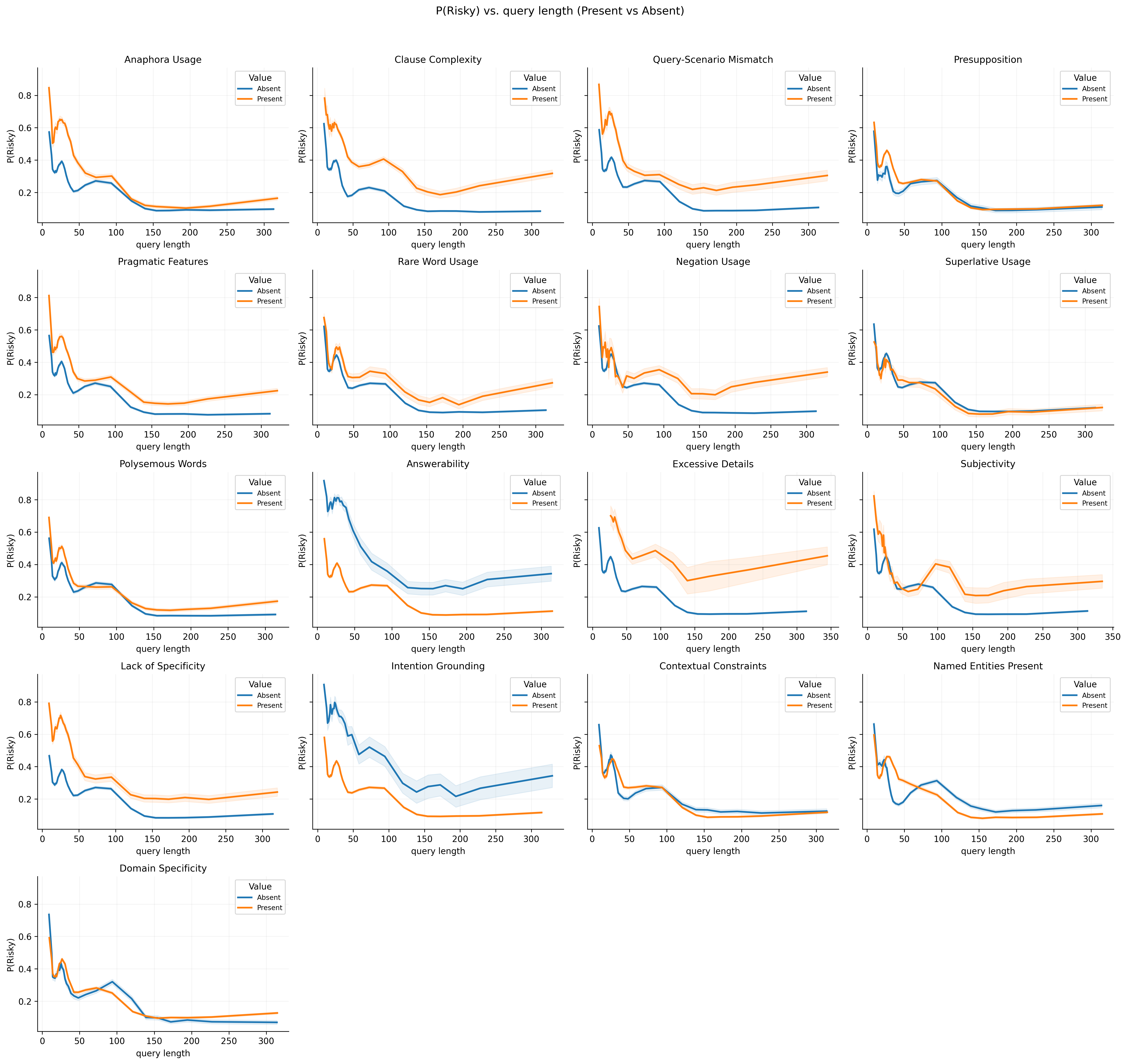}
  \caption{\textbf{Risk vs. query length (Present vs Absent).}
  For each feature, we bin queries into length quantiles and plot the empirical
  probability of the \emph{Risky} label within each bin for \emph{Present} vs
  \emph{Absent}. Shaded bands are binomial CIs. Separation is largest at short
  lengths: features such as \emph{Lack of Specificity} and \emph{Excessive Details}
  increase risk, whereas \emph{Answerability} and \emph{Intention Grounding}
  reduce risk across lengths.}
  \label{fig:len-grid-top}
\end{figure*}


\section{Prompt Templates}
\label{app:prompt-templates}

\begin{promptbox}{Universal Feature Template}
You are an expert linguist. Given a user query, decide whether it exhibits the FEATURE below using the operational rubric.
Return STRUCTURED OUTPUT with fields:
- label: true|false
- rationale: <=2 sentences (short, evidence-based)

FEATURE: {{FEATURE_NAME}}

OPERATIONAL RUBRIC:
- {{RUBRIC_BULLET_1}}
- {{RUBRIC_BULLET_2}}
- {{RUBRIC_BULLET_3}}

EXAMPLES (5-shot; mix of positive and negative):
[E1] FEATURE=Negation Usage; INPUT: Why didn't the test run?
OUTPUT: label=true; rationale="Contains explicit negation (didn't) affecting the main predicate."

[E2] FEATURE=Negation Usage; INPUT: Why did the test run?
OUTPUT: label=false; rationale="No negation markers present."

[E3] FEATURE=Lack of Specificity; INPUT: Tell me about Tesla.
OUTPUT: label=true; rationale="Multiple plausible scopes (company, vehicles, stock) with no constraints."

[E4] FEATURE=Lack of Specificity; INPUT: Summarize Tesla's 2024 Q4 earnings call in 5 bullets.
OUTPUT: label=false; rationale="Time, scope, and format are clearly specified."

[E5] FEATURE=Named Entities Present; INPUT: Did the CDC issue RSV guidance in 2024?
OUTPUT: label=true; rationale="Contains named entity (CDC) and dated reference (2024)."

Now classify the following query.

INPUT:
{{query}}

STRUCTURED OUTPUT:
label=<true|false>; rationale="<two short sentences>"
\end{promptbox}


\begin{promptbox}{Anaphora Usage}
You are an expert linguist. Given a user query, decide whether it exhibits the FEATURE below using the operational rubric.
Return STRUCTURED OUTPUT with fields:
- label: true|false
- rationale: <=2 sentences

FEATURE: Anaphora Usage

OPERATIONAL RUBRIC:
- Contains pronominal/definite references (it/this/that/they/he/she/these/those/that one) with an antecedent not locally introduced.
- Correct interpretation depends on prior discourse or missing antecedent.
- If read in isolation, resolution is unclear or ambiguous.

EXAMPLES (5-shot):
[E1] INPUT: Is he the same person who founded the company?  $\to$ STRUCTURED OUTPUT: label=true; rationale="'he' lacks an antecedent; resolution depends on prior context."
[E2] INPUT: How does this compare to that paper from last year? $\to$ label=true; rationale="'this' and 'that paper' require discourse to resolve."
[E3] INPUT: It was delayed again--when will it ship? $\to$ label=true; rationale="'It' is anaphoric with no antecedent in the query."
[E4] INPUT: Who founded Apple? $\to$ label=false; rationale="No anaphoric expressions; fully self-contained."
[E5] INPUT: Define photosynthesis. $\to$ label=false; rationale="No pronouns or anaphoric references."

Now classify the following query.

INPUT:
{{query}}

STRUCTURED OUTPUT:
label=<true|false>; rationale="<two short sentences>"
\end{promptbox}

\begin{promptbox}{Clause Complexity}
You are an expert linguist. Decide whether the query exhibits the FEATURE below using the rubric. 
Return STRUCTURED OUTPUT with fields {label, rationale (<=2 sentences)}.

FEATURE: Clause Complexity

OPERATIONAL RUBRIC:
- Contains multiple subordinate/relative/conditional clauses.
- Uses subordinators or relativizers (because, although, which/that, if/when/while, even though, so that).
- Meaning would materially change if reduced to a single clause.

EXAMPLES (5-shot):
[E1] INPUT: If the trial succeeds, which regulators will, according to the memo that leaked, approve it first? $\to$ label=true; rationale="Multiple embedded/conditional clauses."
[E2] INPUT: Summarize the study that was published last week, which compared three models. $\to$ label=true; rationale="Relative clauses 'that was published' and 'which compared'."
[E3] INPUT: Although sales fell, margins improved. $\to$ label=true; rationale="Subordinate concessive clause."
[E4] INPUT: Who wrote The Road? $\to$ label=false; rationale="Single simple clause."
[E5] INPUT: Define GDP. $\to$ label=false; rationale="No subordination or embedding."

INPUT:
{{query}}

STRUCTURED OUTPUT:
label=<true|false>; rationale="..."
\end{promptbox}

\begin{promptbox}{Query--Scenario Mismatch}
You are an expert linguist. Decide whether the query is mismatched with the declared scenario.
Return STRUCTURED OUTPUT with fields {label, rationale (<=2 sentences)}.

FEATURE: Query--Scenario Mismatch

OPERATIONAL RUBRIC:
- Requested operation conflicts with SCENARIO (Extractive / Abstractive / Multiple-Choice).
- Expected answer format is incompatible with SCENARIO resources (e.g., asks for "exact span" but no passage; asks to "pick an option" but no options).
- The query presupposes inputs (choices/passage) absent in the scenario.

EXAMPLES (5-shot):
[E1] SCENARIO=Abstractive; INPUT: Extract the exact span containing the date. $\to$ label=true; rationale="Extraction request in Abstractive setting."
[E2] SCENARIO=Multiple-Choice; INPUT: Provide a free-form summary of the article. $\to$ label=true; rationale="Open summary in MC setting."
[E3] SCENARIO=Extractive; INPUT: Choose the correct option (A--D). $\to$ label=true; rationale="MC instruction in Extractive scenario."
[E4] SCENARIO=Abstractive; INPUT: Summarize the passage in three bullets. $\to$ label=false; rationale="Matches Abstractive scenario."
[E5] SCENARIO=Multiple-Choice; INPUT: Select the best answer from the options. $\to$ label=false; rationale="Matches MC scenario."

INPUT:
SCENARIO: {{scenario}}
{{query}}

STRUCTURED OUTPUT:
label=<true|false>; rationale="..."
\end{promptbox}

\begin{promptbox}{Presupposition}
You are an expert linguist. Decide whether the query embeds a nontrivial presupposition.
Return STRUCTURED OUTPUT with {label, rationale<=2 sentences}.

FEATURE: Presupposition

OPERATIONAL RUBRIC:
- Assumes some fact is true (existence/uniqueness/factivity) without evidence in the query.
- Removing the presupposition changes truth conditions (e.g., "When did X stop$\dots$" presupposes X used to$\dots$).
- The assumed fact may be false or unverifiable given typical inputs.

EXAMPLES (5-shot):
[E1] INPUT: When did the CEO admit the fraud? $\to$ label=true; rationale="Presupposes there was a fraud and an admission."
[E2] INPUT: Who is the king of France now? $\to$ label=true; rationale="Presupposes France has a king."
[E3] INPUT: Why did the model fail again? $\to$ label=true; rationale="Presupposes failure occurred previously."
[E4] INPUT: Who wrote Pride and Prejudice? $\to$ label=false; rationale="No hidden assumption beyond existence of the book."
[E5] INPUT: Define inflation. $\to$ label=false; rationale="No presupposed event/state."

INPUT:
{{query}}

STRUCTURED OUTPUT:
label=<true|false>; rationale="..."
\end{promptbox}

\begin{promptbox}{Pragmatic Features}
You are an expert linguist. Decide whether the query relies on pragmatics (implicature, deixis, indirect speech acts).
Return STRUCTURED OUTPUT with {label, rationale<=2 sentences}.

FEATURE: Pragmatic Features

OPERATIONAL RUBRIC:
- Literal form diverges from intended act (e.g., "Can you pass the salt?" = request).
- Meaning depends on deixis ("here", "now", "this time") or shared situational context.
- Interpretation requires implicature/sarcasm/politeness beyond literal semantics.

EXAMPLES (5-shot):
[E1] INPUT: Could you maybe tone that down a bit? $\to$ label=true; rationale="Indirect request, politeness strategy."
[E2] INPUT: It's cold in here. $\to$ label=true; rationale="Likely a request to close window/adjust temp (implicature)."
[E3] INPUT: Is that really how we want to do this? $\to$ label=true; rationale="Rhetorical/indirect suggestion."
[E4] INPUT: What is the capital of Japan? $\to$ label=false; rationale="Literal Q&A."
[E5] INPUT: Define entropy in thermodynamics. $\to$ label=false; rationale="No pragmatic inference required."

INPUT:
{{query}}

STRUCTURED OUTPUT:
label=<true|false>; rationale="..."
\end{promptbox}

\begin{promptbox}{Rare Word Usage}
You are an expert linguist. Decide whether the query uses rare/low-frequency or highly technical terms.
Return STRUCTURED OUTPUT {label, rationale<=2 sentences}.

FEATURE: Rare Word Usage

OPERATIONAL RUBRIC:
- Contains niche jargon or low-frequency lexical items relative to general English.
- Common synonyms exist that would be much more frequent.
- A typical non-expert would flag the term as uncommon.

EXAMPLES (5-shot):
[E1] INPUT: Explain the pathophysiology of rhabdomyolysis. $\to$ label=true; rationale="'rhabdomyolysis' is rare, technical."
[E2] INPUT: Define syzygy in orbital mechanics. $\to$ label=true; rationale="'syzygy' is rare."
[E3] INPUT: What does heteroscedasticity mean? $\to$ label=true; rationale="Technical statistical term."
[E4] INPUT: What is a star? $\to$ label=false; rationale="Common vocabulary."
[E5] INPUT: Who was the first president of the US? $\to$ label=false; rationale="No rare words."

INPUT:
{{query}}

STRUCTURED OUTPUT:
label=<true|false>; rationale="..."
\end{promptbox}

\begin{promptbox}{Negation Usage}
You are an expert linguist. Decide whether the query contains semantic negation.
Return STRUCTURED OUTPUT {label, rationale<=2 sentences}.

FEATURE: Negation Usage

OPERATIONAL RUBRIC:
- Uses explicit negation tokens (not, no, never, without, hardly, scarcely).
- Negation scope changes the truth of the main predicate.
- Negative polarity is central to the request.

EXAMPLES (5-shot):
[E1] INPUT: Which vaccines are not mRNA-based? $\to$ label=true; rationale="Explicit negation 'not' restricting set."
[E2] INPUT: Why didn't the test run? $\to$ label=true; rationale="Negated auxiliary 'didn't'."
[E3] INPUT: Summarize the paper without mentioning formulas. $\to$ label=true; rationale="'without' introduces negation constraint."
[E4] INPUT: Who wrote Hamlet? $\to$ label=false; rationale="No negation."
[E5] INPUT: Define polymerase. $\to$ label=false; rationale="No negation."

INPUT:
{{query}}

STRUCTURED OUTPUT:
label=<true|false>; rationale="..."
\end{promptbox}

\begin{promptbox}{Superlative Usage}
You are an expert linguist. Decide whether the query uses superlatives.
Return STRUCTURED OUTPUT {label, rationale<=2 sentences}.

FEATURE: Superlative Usage

OPERATIONAL RUBRIC:
- Morphological/lexical superlatives (biggest, smallest, "the most/least", "of all").
- Implies an ordering over a set with an extreme endpoint.
- Expects a unique argmax/argmin or tie-breaking criterion.

EXAMPLES (5-shot):
[E1] INPUT: What is the fastest marine mammal? $\to$ label=true; rationale="Superlative 'fastest'."
[E2] INPUT: Which city has the most museums? $\to$ label=true; rationale="'the most' indicates superlative count."
[E3] INPUT: What is the smallest prime number greater than 50? $\to$ label=true; rationale="'smallest' within a constrained set."
[E4] INPUT: Name a city with many museums. $\to$ label=false; rationale="Comparative/quantified, not superlative."
[E5] INPUT: Define prime number. $\to$ label=false; rationale="No superlative."

INPUT:
{{query}}

STRUCTURED OUTPUT:
label=<true|false>; rationale="..."
\end{promptbox}

\begin{promptbox}{Polysemous Words}
You are an expert linguist. Decide whether a key content word is polysemous and under-specified here.
Return STRUCTURED OUTPUT {label, rationale<=2 sentences}.

FEATURE: Polysemous Words

OPERATIONAL RUBRIC:
- A salient word has multiple distinct senses (bank, cell, Java, Mercury).
- Local context does not disambiguate the intended sense.
- Different senses would change the answer.

EXAMPLES (5-shot):
[E1] INPUT: How do I open a new account at the bank? $\to$ label=false; rationale="Context favors financial institution."
[E2] INPUT: What is the weather like in Java? $\to$ label=true; rationale="Could be island or language; under-specified."
[E3] INPUT: Describe the function of a cell. $\to$ label=true; rationale="Could be biological cell or prison cell."
[E4] INPUT: Mercury's orbital period is what? $\to$ label=true; rationale="Planet vs. element; ambiguous."
[E5] INPUT: Who wrote The Hobbit? $\to$ label=false; rationale="No polysemous ambiguity."

INPUT:
{{query}}

STRUCTURED OUTPUT:
label=<true|false>; rationale="..."
\end{promptbox}

\begin{promptbox}{Answerability}
You are an expert linguist. Decide whether the query is answerable on the basis of provided/commonly-known information (not speculation).
Return STRUCTURED OUTPUT {label, rationale<=2 sentences}.

FEATURE: Answerability

OPERATIONAL RUBRIC:
- Has a verifiable answer given supplied context or widely-known facts.
- Not opinion-based, rhetorical, or forecasting without data.
- Does not require time-varying external info unless included.

EXAMPLES (5-shot):
[E1] INPUT: Who wrote The Road? $\to$ label=true; rationale="Single verifiable fact (Cormac McCarthy)."
[E2] INPUT: What is $17\times19$? $\to$ label=true; rationale="Deterministic computation."
[E3] INPUT: Will Stock X crash next month? $\to$ label=false; rationale="Speculative forecasting."
[E4] INPUT: Should I move to New York? $\to$ label=false; rationale="Subjective; no criteria."
[E5] INPUT: Is there life on Europa? $\to$ label=false; rationale="Unknown; not currently verifiable."

INPUT:
{{query}}

STRUCTURED OUTPUT:
label=<true|false>; rationale="..."
\end{promptbox}

\begin{promptbox}{Excessive Details}
You are an expert linguist. Decide whether the query includes extraneous details not needed to answer it.
Return STRUCTURED OUTPUT {label, rationale<=2 sentences}.

FEATURE: Excessive Details

OPERATIONAL RUBRIC:
- Contains descriptive asides that do not constrain the answer.
- Removing them would not change the target operation or output.
- Details distract or broaden scope without adding specificity.

EXAMPLES (5-shot):
[E1] INPUT: In my blue notebook from last summer's trip to Italy, can you define mitosis? $\to$ label=true; rationale="Notebook/trip details irrelevant to defining mitosis."
[E2] INPUT: Please, given my favorite mug and desk plant, what is $12\times8$? $\to$ label=true; rationale="Superfluous objects unrelated to arithmetic."
[E3] INPUT: When did WWI begin? $\to$ label=false; rationale="No extra details."
[E4] INPUT: Summarize this article in 3 bullets. $\to$ label=false; rationale="No extraneous info."
[E5] INPUT: What is the boiling point of water at sea level? $\to$ label=false; rationale="All details are relevant."

INPUT:
{{query}}

STRUCTURED OUTPUT:
label=<true|false>; rationale="..."
\end{promptbox}

\begin{promptbox}{Subjectivity}
You are an expert linguist. Decide whether the query requests a subjective judgment or preference.
Return STRUCTURED OUTPUT {label, rationale<=2 sentences}.

FEATURE: Subjectivity

OPERATIONAL RUBRIC:
- Invites personal taste/value judgment (best, beautiful, should, worth) without criteria.
- No objective rubric is provided to adjudicate correctness.
- Output depends on preferences rather than evidence.

EXAMPLES (5-shot):
[E1] INPUT: Which smartphone is the best right now? $\to$ label=true; rationale="'best' without criteria is subjective."
[E2] INPUT: Should I learn Rust or Go? $\to$ label=true; rationale="Advisory preference question."
[E3] INPUT: Is modern art good? $\to$ label=true; rationale="Value judgment."
[E4] INPUT: What's the battery capacity of iPhone 13? $\to$ label=false; rationale="Objective spec."
[E5] INPUT: Define convolution. $\to$ label=false; rationale="Objective definition."

INPUT:
{{query}}

STRUCTURED OUTPUT:
label=<true|false>; rationale="..."
\end{promptbox}

\begin{promptbox}{Lack of Specificity}
You are an expert linguist. Decide whether the query is under-specified.
Return STRUCTURED OUTPUT {label, rationale<=2 sentences}.

FEATURE: Lack of Specificity

OPERATIONAL RUBRIC:
- Missing disambiguating constraints (time/place/entity/scope).
- Multiple plausible interpretations; no tie-breaker.
- Task intent or output format is underspecified.

EXAMPLES (5-shot):
[E1] INPUT: Tell me about Tesla. $\to$ label=true; rationale="Company vs. cars vs. stock; scope unclear."
[E2] INPUT: Compare the models. $\to$ label=true; rationale="Which models? No domain or criteria."
[E3] INPUT: What happened yesterday? $\to$ label=true; rationale="No topic or domain given."
[E4] INPUT: Summarize Tesla's 2024 Q4 earnings call in 5 bullets. $\to$ label=false; rationale="Time, domain, and format specified."
[E5] INPUT: Extract the date of publication from the abstract. $\to$ label=false; rationale="Clear operation and target."

INPUT:
{{query}}

STRUCTURED OUTPUT:
label=<true|false>; rationale="..."
\end{promptbox}

\begin{promptbox}{Intention Grounding}
You are an expert linguist. Decide whether the user's intended operation is explicit.
Return STRUCTURED OUTPUT {label, rationale<=2 sentences}.

FEATURE: Intention Grounding

OPERATIONAL RUBRIC:
- Verb makes the operation clear (summarize, compare, extract, classify, translate).
- Expected output form is inferable (bullets, short answer, definition).
- Operation applies to supplied or implied content.

EXAMPLES (5-shot):
[E1] INPUT: Summarize the article in three bullets. $\to$ label=true; rationale="Clear directive and format."
[E2] INPUT: Extract the chemical formula from the passage. $\to$ label=true; rationale="Unambiguous extraction task."
[E3] INPUT: Compare Model A and Model B on latency and cost. $\to$ label=true; rationale="Operation and criteria stated."
[E4] INPUT: Java? $\to$ label=false; rationale="No operation specified."
[E5] INPUT: Tell me about space. $\to$ label=false; rationale="Vague goal, no operation."

INPUT:
{{query}}

STRUCTURED OUTPUT:
label=<true|false>; rationale="..."
\end{promptbox}

\begin{promptbox}{Contextual Constraints}
You are an expert linguist. Decide whether the query includes explicit constraints that narrow scope.
Return STRUCTURED OUTPUT {label, rationale<=2 sentences}.

FEATURE: Contextual Constraints

OPERATIONAL RUBRIC:
- Names time, location, population, or conditions that meaningfully narrow the answer.
- Constraints are integral to fulfilling the request.
- Removing constraints would broaden or change the target.

EXAMPLES (5-shot):
[E1] INPUT: List three causes of inflation in the US during 2022. $\to$ label=true; rationale="Time and location constraints."
[E2] INPUT: Summarize EU AI Act obligations for SMEs only. $\to$ label=true; rationale="Jurisdiction and population constraints."
[E3] INPUT: Give NYC subway delays after 10pm. $\to$ label=true; rationale="Location and time constraints."
[E4] INPUT: Define inflation. $\to$ label=false; rationale="No constraints."
[E5] INPUT: Summarize the article. $\to$ label=false; rationale="No narrowing conditions."

INPUT:
{{query}}

STRUCTURED OUTPUT:
label=<true|false>; rationale="..."
\end{promptbox}

\begin{promptbox}{Named Entities Present}
You are an expert linguist. Decide whether the query includes named entities (proper names).
Return STRUCTURED OUTPUT {label, rationale<=2 sentences}.

FEATURE: Named Entities Present

OPERATIONAL RUBRIC:
- Contains proper names (persons, orgs, places, products, works, dates).
- Entities are pivotal to resolving the query.
- Generic categories alone (city, company) do not count as named entities.

EXAMPLES (5-shot):
[E1] INPUT: Did Sundar Pichai announce Gemini in 2023? $\to$ label=true; rationale="Person and product names; year."
[E2] INPUT: What did the CDC advise about RSV in 2024? $\to$ label=true; rationale="Org and year."
[E3] INPUT: When did World War I begin? $\to$ label=true; rationale="Named historical event."
[E4] INPUT: Who wrote that book? $\to$ label=false; rationale="No explicit names given."
[E5] INPUT: Define a balanced tree. $\to$ label=false; rationale="No proper names."

INPUT:
{{query}}

STRUCTURED OUTPUT:
label=<true|false>; rationale="..."
\end{promptbox}

\begin{promptbox}{Domain Specificity}
You are an expert linguist. Decide whether the query is specialized to a technical/professional domain.
Return STRUCTURED OUTPUT {label, rationale<=2 sentences}.

FEATURE: Domain Specificity

OPERATIONAL RUBRIC:
- Requires discipline-specific knowledge/terminology (law, medicine, finance, ML, biology, etc.).
- A layperson would likely consult an expert/reference.
- Uses domain conventions or statutory/technical terms.

EXAMPLES (5-shot):
[E1] INPUT: Interpret Section 230 immunity in the context of platform moderation. $\to$ label=true; rationale="Legal doctrine."
[E2] INPUT: Differentiate Type I vs. Type II error and discuss $\alpha$ control. $\to$ label=true; rationale="Statistical terminology."
[E3] INPUT: Explain CRISPR-Cas9 off-target effects. $\to$ label=true; rationale="Molecular biology."
[E4] INPUT: What is a triangle? $\to$ label=false; rationale="Elementary concept."
[E5] INPUT: Who was Ada Lovelace? $\to$ label=false; rationale="General knowledge biography."

INPUT:
{{query}}

STRUCTURED OUTPUT:
label=<true|false>; rationale="..."
\end{promptbox}

\begin{promptbox}{Paraphrase Generation ($K \leq 6$)}
You are a careful rewriter. Generate up to 6 paraphrases that preserve the original meaning exactly while avoiding any added or lost constraints.
Return STRUCTURED OUTPUT with fields:
- paraphrases: list of strings (length $1\dots6$)

RULES:
- Preserve all factual content, entities, temporal/locational constraints, and requested operation.
- Do not simplify away constraints or add new ones.
- Keep reading level and tone similar; avoid stylistic extremes.
- Each paraphrase must be self-contained and faithful.

EXAMPLES (5-shot):
[E1]
INPUT: Summarize the article in three bullets.
OUTPUT:
paraphrases=[
  "Provide a three-bullet summary of the article.",
  "Summarize the article using exactly three bullet points.",
  "Give a three-point bulleted summary of the article."
]

[E2]
INPUT: Extract the publication date from the abstract.
OUTPUT:
paraphrases=[
  "Identify the publication date contained in the abstract.",
  "From the abstract, extract the date of publication."
]

[E3]
INPUT: Compare Model A and Model B on latency and cost.
OUTPUT:
paraphrases=[
  "Compare Model A with Model B in terms of latency and cost.",
  "Evaluate Model A versus Model B focusing on latency and cost."
]

[E4]
INPUT: List three causes of inflation in the US during 2022.
OUTPUT:
paraphrases=[
  "List three causes of U.S. inflation in 2022.",
  "Provide three factors that drove inflation in the United States during 2022."
]

[E5]
INPUT: Translate the sentence into Spanish without adding explanations.
OUTPUT:
paraphrases=[
  "Translate the sentence to Spanish with no additional explanation.",
  "Render the sentence in Spanish only; do not include any commentary."
]

INPUT:
{{query}}

STRUCTURED OUTPUT:
paraphrases=["...", "..."]
\end{promptbox}

\begin{promptbox}{Hallucination LLM-Judge}
You are an impartial grader. Decide whether the model's answer is correct for the given query and gold reference.
Return STRUCTURED OUTPUT with fields:
- correct: true|false
- rationale: <=2 sentences (concise, cite the decisive mismatch or match)

CRITERIA:
- Accept semantic equivalence to the gold even if phrased differently.
- Penalize invented facts, contradictions, or unsupported specifics.
- If gold is a set, accept any member as correct.
- If the gold is not applicable or missing, mark correct=false and explain.

EXAMPLES (5-shot):
[E1]
INPUT:
query: Who wrote "The Road"?
model_answer: Cormac McCarthy.
gold: Cormac McCarthy
OUTPUT:
correct=true; rationale="Exact match to the reference author."

[E2]
INPUT:
query: What is the boiling point of water at sea level?
model_answer: $90^\circ$C.
gold: $100^\circ$C
OUTPUT:
correct=false; rationale="Numerical value contradicts the reference ($90\not=100$)."

[E3]
INPUT:
query: Name one prime number greater than 10.
model_answer: 13.
gold: {11, 13, 17, 19, ...}
OUTPUT:
correct=true; rationale="Answer (13) is a valid member of the acceptable set."

[E4]
INPUT:
query: Define photosynthesis.
model_answer: It is the process by which plants convert light into chemical energy, producing glucose and oxygen from carbon dioxide and water.
gold: Process converting light energy into chemical energy, producing glucose and oxygen from CO$_2$ and water.
OUTPUT:
correct=true; rationale="Semantically equivalent definition."

[E5]
INPUT:
query: Who is the current king of France?
model_answer: Louis XX.
gold: No current king of France.
OUTPUT:
correct=false; rationale="Asserts a non-existent monarch; contradicts the reference."

Now grade the following example.

INPUT:
query: {{query}}
model_answer: {{answer}}
gold: {{gold}}

STRUCTURED OUTPUT:
correct=<true|false>; rationale="<two short sentences>"
\end{promptbox}


\begin{figure*}[t]
    \centering
    \includegraphics[width=0.9\textwidth]{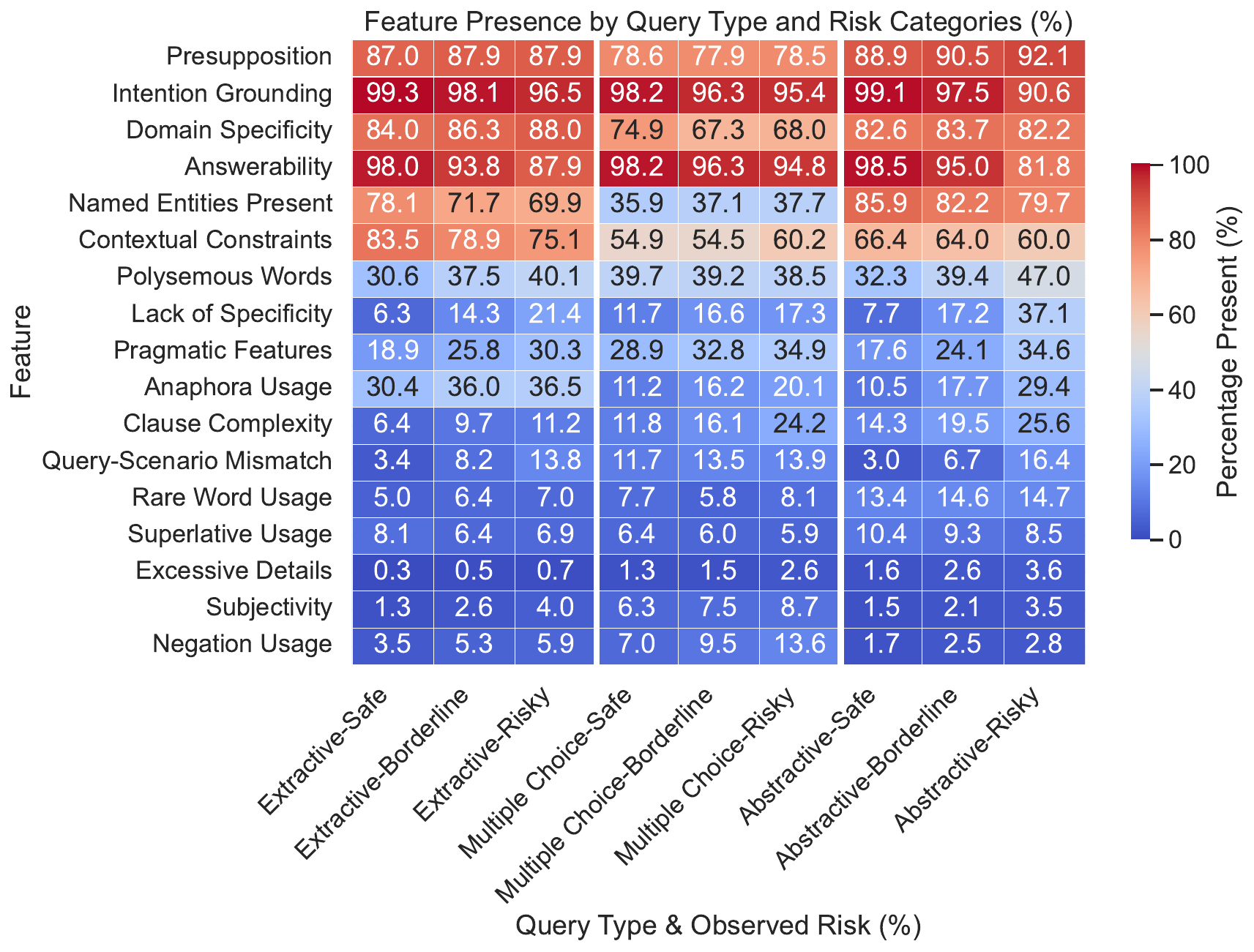}
    \caption{
    Heatmap of the percentage of queries exhibiting each binary linguistic feature, grouped by query type (extractive, multiple choice, abstractive) and categorized by observed risk level (\emph{Safe}, \emph{Borderline}, \emph{Risky}). Warmer colors (reds) indicate higher prevalence of a feature, while cooler colors (blues) indicate lower prevalence. Several features (\emph{lack of specificity}, \emph{clause complexity}, \emph{polysemous words}) increase most prominently from \emph{Safe} to \emph{Risky}, showing a clear monotonic rise in prevalence across risk categories. In contrast, \emph{answerability} and \emph{intention grounding} decrease steadily, and certain features (\emph{domain specificity} and \emph{contextual constraints}) display opposite trends across different query types.
    }
    \label{fig:descriptive-analysis-heatmap}
\end{figure*}

\begin{figure*}[t]
  \centering
  \includegraphics[width=\linewidth]{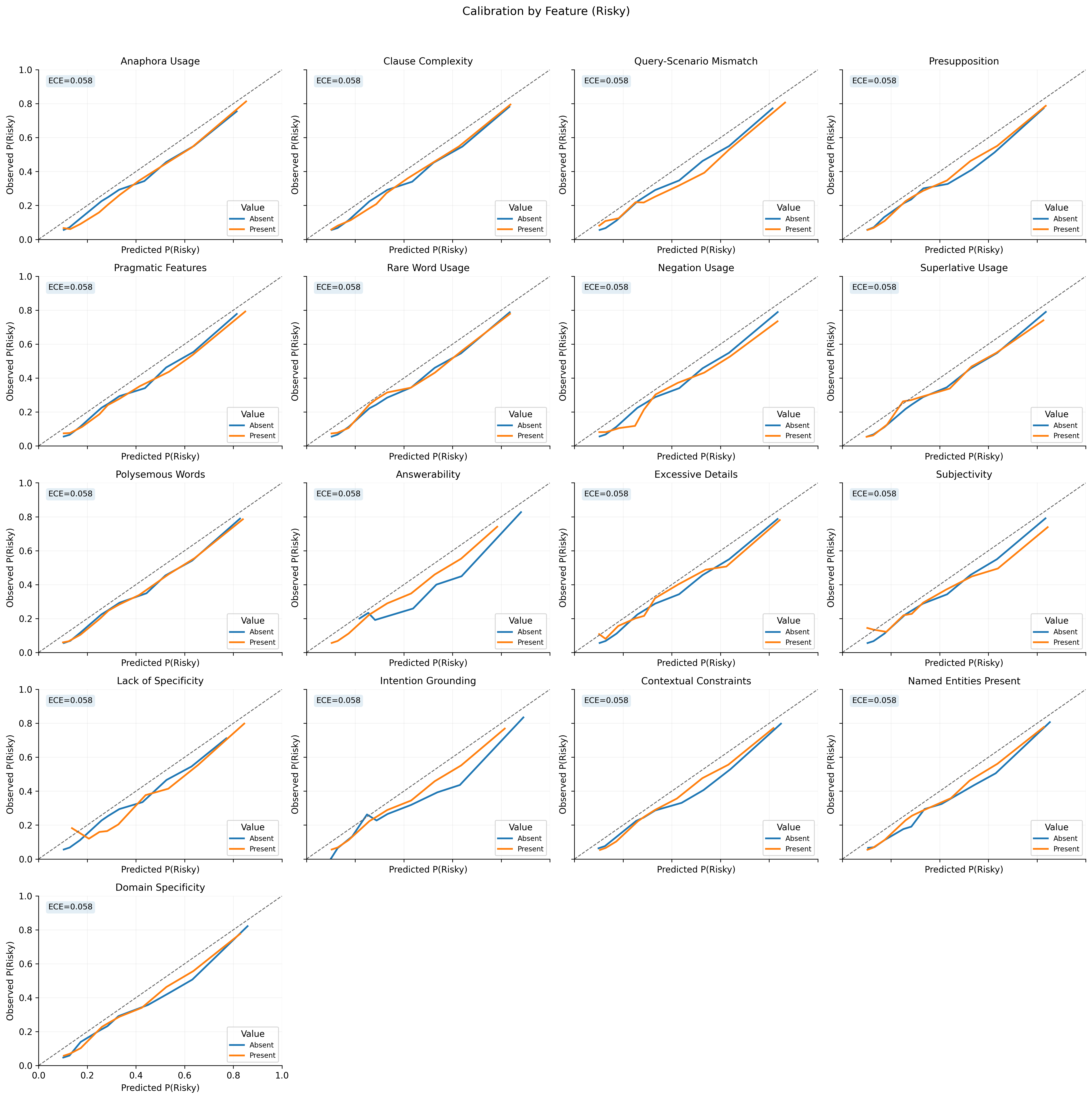}
  \caption{\textbf{Reliability by feature.}
  Binned reliability curves for predicted $P(\text{Risky})$ when a feature is
  \emph{Present} vs \emph{Absent}. Dashed line is perfect calibration.
  An overall ECE is reported per panel (10 equal-mass bins).
  Calibration across strata. We show reliability curves stratified by feature presence. The model is reasonably calibrated across strata (ECE $\approx$ 0.05–0.06). Importantly, the \emph{direction} of miscalibration does not reverse between Present/Absent strata for the dominant features (e.g., Answerability, Lack of Specificity), supporting that the feature effects observed in the ECDFs translate to well-behaved risk scores rather than artifacts of calibration.
  }
  \label{fig:calib-all}
\end{figure*}

\begin{figure*}[t]
  \centering
  \includegraphics[width=\linewidth]{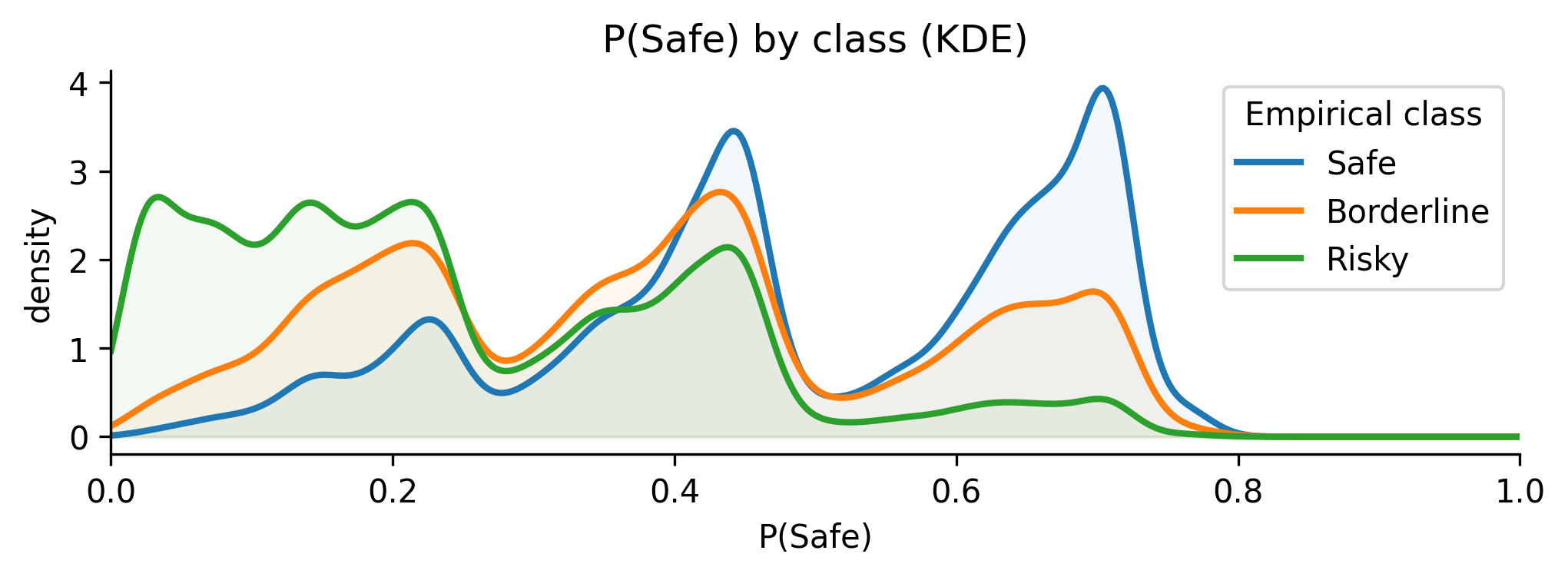}
  
  \includegraphics[width=\linewidth]{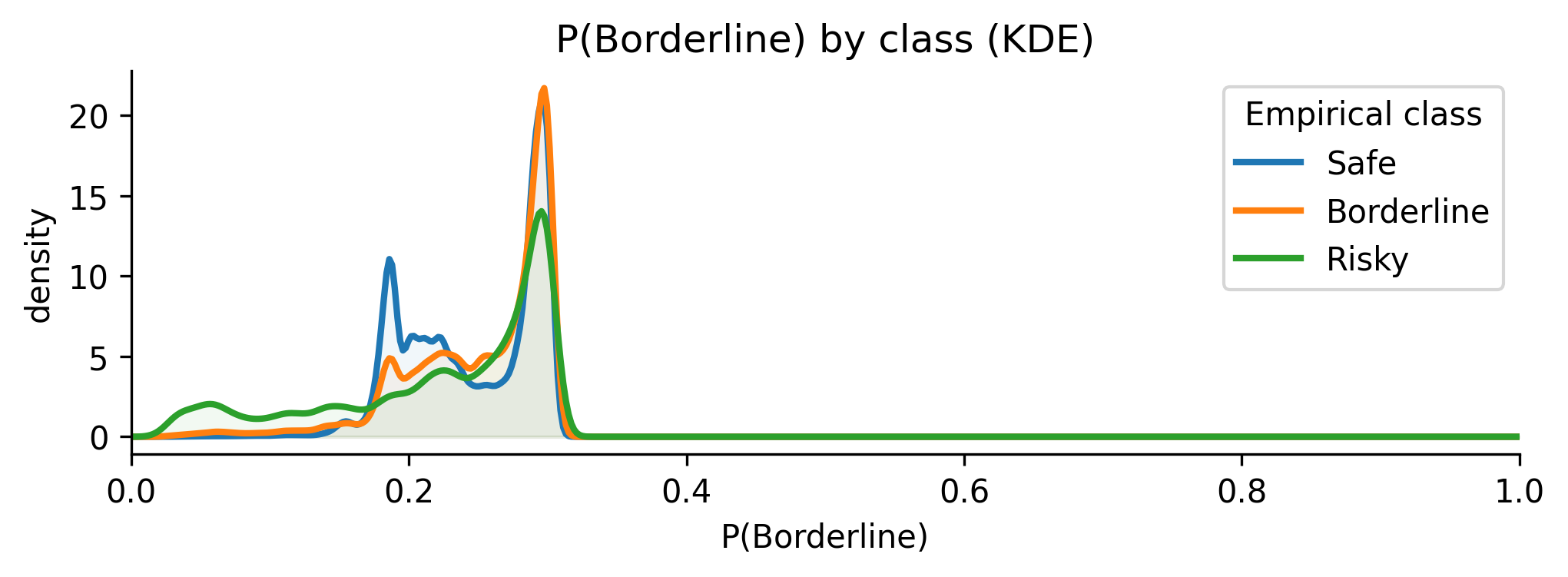}
  
  \includegraphics[width=\linewidth]{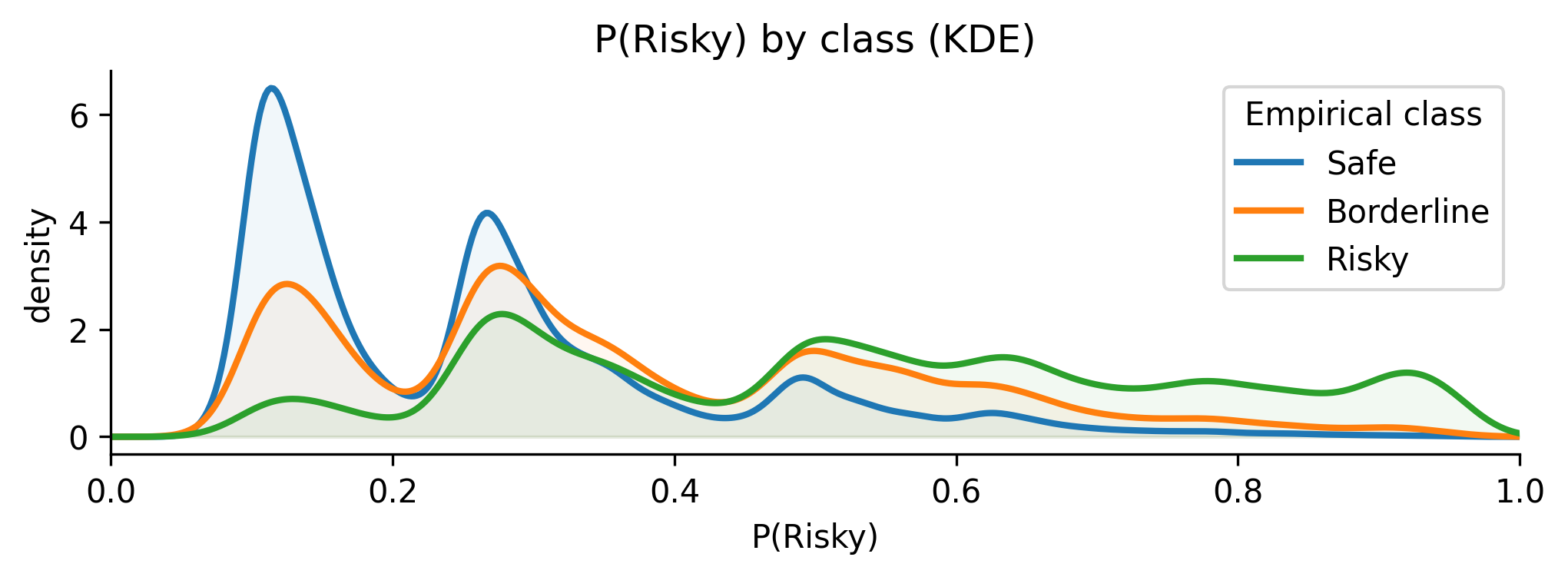}
  \caption{\textbf{Per-class probability KDEs.}
  KDEs of model-predicted $P(\text{Safe})$, $P(\text{Borderline})$,
  and $P(\text{Risky})$ grouped by empirical class labels.}
  \label{fig:kde-per-class}
\end{figure*}

\begin{table*}[ht]
\small
\centering
\renewcommand{\arraystretch}{1.5}
\setlength{\tabcolsep}{8pt}
\begin{tabular}{clp{5cm}p{4.5cm}}
\toprule
{} & \textbf{Feature} & \textbf{Definition} & \textbf{Example} \\
\midrule

\multirow{10}{*}{\rotatebox{90}{\textbf{Structural}}} 
    & Query and Context Length & Total number of tokens in the query (and context, when applicable) & "How does reinforcement learning work?" (6 tokens) \\
    & Anaphoric Reference & Presence of pronouns or references requiring external context. & "What about that one?" (Unclear reference) \\
    & Clause Complexity & Measures the presence of multiple subordinate clauses & "While I was walking home, I saw a cat that looked just like my friend's." \\
    & Dependency Tree Depth & Depth of the query's syntactic dependency tree. & "Describe the structure of a sentence that contains multiple levels of embedding." (Depth: 7) \\
    & Parse Tree Height & Height of the parse tree, providing a secondary measure of syntactic complexity. & "Analyze a sentence with nested relative clauses." (Height: 4) \\
\midrule

\multirow{7}{*}{\rotatebox{90}{\textbf{Scenario-Based}}} 
    & Query Type & Extractive, Multiple Choice, or Abstractive. & "Summarize the latest economic report." (Requires retrieval) \\
    & Query Scenario Mismatch & Mismatch between the query's intended output and its actual structure. & "List all prime numbers" (Infeasible output expectation) \\
    & Presupposition & Unstated assumptions embedded in the query. & "Who is the musician that developed neural networks?" (Assumes such a musician exists) \\
    & Pragmatics & Captures context-dependent meanings beyond literal interpretation. & "Can you pass the salt?" (A request, not a literal ability) \\
\midrule

\multirow{7}{*}{\rotatebox{90}{\textbf{Lexical}}}
    & Word Rarity & Use of rare or niche terminology. & "What are the ramifications of quantum decoherence?" (Uses low-frequency terms) \\
    & Negation Usage & Presence of negation words (\textit{not}, \textit{never}). & "Is it not possible to do this?" \\
    & Superlatives & Detection of superlative expressions (\textit{biggest}, \textit{fastest}). & "What is the fastest algorithm?" \\
    & Polysemy & Presence of ambiguous words with multiple related meanings. & "Explain how a bank operates." (Ambiguity: financial institution vs. riverbank) \\
\midrule

\multirow{7}{*}{\rotatebox{90}{\textbf{Stylistic}}} 
    & Answerability & Assesses whether the query has a verifiable answer. & "What is the exact number of galaxies?" (Unanswerable) \\
    & Excessive Details & Evaluates whether a query is overloaded with information, potentially distracting the model. & "Can you explain how convolutional neural networks work, including all mathematical formulas?" \\
    & Subjectivity & Detects the degree of opinion or personal bias in the query. & "What is the best programming language?" \\
    & Lack of Specificity & Assesses the breadth or vagueness of a query. & "Tell me about history." (Too broad) \\
\midrule

\multirow{8}{*}{\rotatebox{90}{\textbf{Semantic}} }
    & Intention Grounding & Evaluates how clearly the query's purpose is expressed. & "How does reinforcement learning optimize control in robotics?" (Clear intent) \\
    & Contextual Constraints & Identifies explicit constraints (time, location, conditions) provided in the query. & "What was the inflation rate in the US in 2023?" \\
    & Named Entity Presence & Checks for the inclusion of verifiable named entities. & "Who founded OpenAI?" \\
    & Domain Specificity & Determines whether the query belongs to a specialized domain (e.g., finance, law). & "What are the legal implications of the GDPR ruling?" \\
\bottomrule
\end{tabular}
\caption{Summary of our feature categories, definitions, and examples (See Section~\ref{sec:method})}
\label{tab:feature_dimensions}
\end{table*}

\begin{table*}[ht]
\centering
\small
\renewcommand{\arraystretch}{1.5}
\setlength{\tabcolsep}{6pt}
\begin{tabular}{p{1.5cm}cp{5.5cm}p{5.5cm}}
\toprule
\textbf{Feature} & \textbf{Presence} & \textbf{Question} & \textbf{Chain of Thought} \\
\midrule
Anaphora Usage & \cmark & Who was the guitarist for the English Rock band who Terry Kirkbride performed live in the studio with? & The question contains an anaphoric reference (`the English Rock band') without clear contextual information. \\
                   & \xmark & Isotopes are named for their number of protons plus what? & The question does not contain anaphoric references; it is a straightforward scientific inquiry. \\
\midrule
Clause Complexity & \cmark & During evolution, something happened to increase the size of what organ in humans, relative to that of the chimpanzee? & The query has multiple clauses, increasing its complexity. \\
                  & \xmark & What do some animals do to adjust to hot temperatures? & The question is simple, consisting of a single clause. \\
\midrule
Query-Scenario Mismatch & \cmark & What type of forested areas can be found on the highest terrace? & The query asks about `forested areas,' but without a specific location or context, creating a mismatch. \\
                        & \xmark & What date in 2009 saw the heaviest UK snowfall since 1991? & The question has a direct and valid scenario, asking for a factual historical date. \\
\midrule
Presupposition & \cmark & Central America's Panama seceded from which country in 1903? & The question presupposes that Panama seceded from a specific country in 1903. \\
               & \xmark & What is the scientific name of the true creature featured in ``Creature from the Black Lagoon''? & The question does not assume any prior knowledge; it is a straightforward request for a name. \\
\midrule
Pragmatic Features & \cmark & Where did this pattern come from? & The meaning of `this pattern' relies on prior discourse, making pragmatics necessary. \\
                   & \xmark & What is the name of plant-like protists? & The question does not rely on pragmatics; it seeks a factual term. \\
\midrule
Rare Word Usage & \cmark & Where in the human body can you find the Trapezium bone? & The term `Trapezium' is a less commonly known anatomical term. \\
            & \xmark & What is an organism at the top of the food chain called? & The phrase `apex predator' is well known and lacks rare words. \\
\midrule
Negation Usage & \cmark & Which is not an inherited trait in humans? & The presence of `not' reverses the expectation of the query. \\
               & \xmark & Along with Walt Disney, who created Oswald the Lucky Rabbit? & The question is affirmative without negation. \\
\midrule
Superlative Usage & \cmark & What is the first stage of cellular respiration? & The word `first' introduces a superlative comparison. \\
                  & \xmark & Which river forms a natural border between Argentina and Uruguay? & No superlative forms are present in the query. \\
\midrule
Named Entities Present & \cmark & What borough are the neighborhood of Chelsea and the office building, 10 Hudson Yards, both a part of? & Named entities include `Chelsea' and `10 Hudson Yards.' \\
                      & \xmark & Some plants can detect increased levels of what when reflected from leaves of encroaching neighbors? & No specific named entities are present in the query. \\

\bottomrule
\end{tabular}
\caption{Representative queries illustrating the presence and absence of selected linguistic features, with accompanying chain-of-thought explanations (Part 1).}
\label{tab:feature_presence_1}
\end{table*}

\begin{table*}[ht]
\centering
\small
\renewcommand{\arraystretch}{1.5}
\setlength{\tabcolsep}{6pt}
\begin{tabular}{p{1.5cm}cp{5.5cm}p{5.5cm}}
\toprule
\textbf{Feature} & \textbf{Presence} & \textbf{Question} & \textbf{Chain of Thought} \\
\midrule
Polysemous Words & \cmark & Who supervised the sting operation that implicated Evelyn Dawn Knight? & The word `supervised' could have different meanings but in this context refers to oversight. \\
                 & \xmark & Which string instrument often played the basso continuo parts? & The terms `string instrument' and `basso continuo' are not polysemous in this context. \\
\midrule
Subjectivity & \cmark & What is a criticism of other streaming services? & The query invites subjective responses based on personal opinions. \\
             & \xmark & What is the second book in the Harry Potter series? & The question is factual and does not involve subjectivity. \\
\midrule
Answerability & \cmark & How long was Warsaw occupied by Germany? & The question can be answered based on explicit historical data. \\
              & \xmark & Beyoncé would take a break from music in which year? & The event may not have a definitive, verifiable answer. \\
\midrule
Excessive Details & \cmark & SkyWest Airlines is a North American airline owned by SkyWest, Inc. and headquartered in which city in Utah, U.S., it flies as SkyWest Airlines in a partnership with Alaska Airlines? & The question includes excessive details about partnerships that are unnecessary for identifying the headquarters. \\
                 & \xmark & What is giving birth to dogs called? & The question is concise and does not contain excessive information. \\
\midrule
Domain Specificity & \cmark & What is the term for a series of biochemical reactions by which an organism converts a given reactant to a specific end product? & The question is highly specific to biochemistry. \\
                   & \xmark & Fado is a type of folk music found in which country? & The question is not highly specialized; it relates to general cultural knowledge. \\
\midrule
Lack of Specificity & \cmark & What division is the Canadian Army Doctrine of? & The query lacks clarity in defining what is meant by `division.' \\
                    & \xmark & Winchester was the capital of which Anglo Saxon kingdom? & The question is specific in its historical context. \\
\midrule
Intention Grounding & \cmark & Which of the two mines, Discovery Mine or Big Dan Mine, produced more gold? & The question is well-grounded in intent by seeking a clear comparison. \\
                    & \xmark & What are the two blocks of Catalan? & The intention is unclear due to the ambiguity of `blocks.' \\
\midrule
Contextual Constraints & \cmark & Which is the least densely populated county of England? & The question is constrained to a specific geographical location. \\
                       & \xmark & Who was the lyricist partner of Richard Rogers prior to Oscar Hammerstein? & No explicit constraints limit the question's scope. \\
\bottomrule
\end{tabular}
\caption{Representative queries illustrating the presence and absence of selected linguistic features, with accompanying chain-of-thought explanations (Part 2).}
\label{tab:feature_presence_2}
\end{table*}

\end{document}